\documentclass{article}

\PassOptionsToPackage{numbers, compress}{natbib}
\usepackage[final]{neurips_2025}

\usepackage[table,xcdraw,dvipsnames]{xcolor}
\usepackage[utf8]{inputenc} % allow utf-8 input
\usepackage[T1]{fontenc}    % use 8-bit T1 fonts
\usepackage[hidelinks]{hyperref}       % hyperlinks
\usepackage{url}            % simple URL typesetting
\usepackage{booktabs}       % professional-quality tables
\usepackage{amsfonts}       % blackboard math symbols
\usepackage{nicefrac}       % compact symbols for 1/2, etc.
\usepackage{microtype}      % microtypography
\usepackage{xcolor}         % colors
\usepackage{enumitem}
\usepackage{wrapfig}
\usepackage{amsmath, amsthm, amssymb}
\usepackage{adjustbox}
\usepackage{multirow}
\usepackage{textcomp}
\usepackage{pifont}
\usepackage{subcaption}
\usepackage[font=small, tableposition=top]{caption}
\usepackage[misc]{ifsym}
\usepackage{floatrow}
\usepackage{bm}
\usepackage{array}
\usepackage{arydshln}
\usepackage{verbatim}
\usepackage{comment}
\usepackage{listings}
\usepackage{xspace}
\usepackage[subtle]{savetrees}
\def\eqref#1{equation~\ref{#1}}
\def\1{\bm{1}}

\def\vg{{\bm{g}}}

\def\vk{{\bm{k}}}

\def\vq{{\bm{q}}}

\def\vv{{\bm{v}}}

\def\vy{{\bm{y}}}
\def\vz{{\bm{z}}}

\def\mK{{\bm{K}}}

\def\mV{{\bm{V}}}

\DeclareMathAlphabet{\mathsfit}{\encodingdefault}{\sfdefault}{m}{sl}
\SetMathAlphabet{\mathsfit}{bold}{\encodingdefault}{\sfdefault}{bx}{n}

\newcommand{\rat}{\texttt{RAT}\xspace}
\newcommand{\swa}{SWA\xspace}
\newcommand{\ratswa}{\texttt{RAT-SWA}\xspace}
\newcommand{\rnn}{RNN\xspace}

\newcommand{\attn}{attention\xspace}

\newcommand{\ratl}[1]{\texttt{RAT(L=#1)}\xspace}
\newcommand{\ratlswa}[1]{\texttt{RAT(L=#1)-SWA}\xspace}
\newcommand{\attnswa}{Attention-SWA\xspace}

\newcolumntype{H}{>{\setbox0=\hbox\bgroup}c<{\egroup}@{}}

\def\equationautorefname~#1\null{Eq.~(#1)\null}

\definecolor{darkorange}{RGB}{255,140,0}
\colorlet{ours}{darkorange!60}

\definecolor{lightskyblue}{RGB}{135,206,250}
\colorlet{baseline}{lightskyblue!60}

\definecolor{ours-gray}{gray}{0.82}
\definecolor{baseline-gray}{gray}{0.94}

\newcommand{\ratrowall}{\rowcolor{ours-gray}\rule{0pt}{2.5ex}}

\newcommand\ratlogo{\raisebox{-8pt}{\includegraphics[width=1.5em]{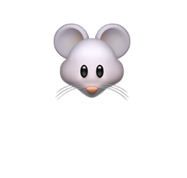}}}

\title{\rat\ratlogo: Bridging RNN Efficiency and Attention Accuracy via Chunk-based Sequence Modeling}

\author{
\textbf{Xiuying Wei\textsuperscript{1}\thanks{Correspondence to \texttt{xiuying.wei@epfl.ch}.} , Anunay Yadav\textsuperscript{1}, Razvan Pascanu\textsuperscript{2}, Caglar Gulcehre\textsuperscript{1}} \\
\textsuperscript{1}CLAIRE, EPFL  %, Lausanne, Switzerland \\
\textsuperscript{2}Google DeepMind \\
}

\begin{document}

\maketitle

\begin{abstract}
Transformers have become the cornerstone of modern large-scale language models, but their reliance on softmax attention poses a computational bottleneck at both training and inference. Recurrent models offer high efficiency, but compressing the full sequence into a fixed-size and holistic representation can suffer from memory degradation in long contexts and limit fine-grained retrieval. To address this, we propose \rat, an intermediate design that bridges the efficiency of RNNs and capacity of attention. \rat partitions the input into chunks, applies recurrence within each chunk for local dependencies, and softmax-based attention across chunks for long-range interactions. This design mitigates memory degradation and enables direct access to distant tokens, while retaining computational efficiency. Empirically, with a chunk size of 16, the \rat block achieves a \(7\times\) improvement in training speed for 100K sequence length and \(9\times\) in generation at the 4K position, while maintaining similar performance compared to standard attention. We demonstrate this by training 1.3B parameter models from scratch and performing large-scale evaluations, including short- and long-context benchmarks, as well as supervised fine-tuning~(SFT). We further propose a hybrid architecture that interleaves \rat with local attention. By combining efficient long-range modeling with strong local interactions, this hybrid design not only improves inference speed and reduces cache memory usage, but also consistently enhances performance and shows the overall best results. Code is available at \url{https://github.com/CLAIRE-Labo/RAT}.
\end{abstract}

\section{Introduction}
Language modeling has long been dominated by Transformer-based architectures due to their strong performance across a wide range of tasks. However, their reliance on full self-attention~\citep{attention} results in quadratic time and memory complexity with respect to sequence length, which limits scalability in long-context processing. This limitation has motivated a wave of recent efforts to revisit recurrent models or propose novel linear recurrent models such as state space models, linear attention methods~\citep{mamba,mamba2,linear_attention,gateddeltanet,xlstm,griffin,lru}.

By comparing these architectures, we observe a key difference between recurrent models and self-attention. Recurrent approaches compress the full sequence history into fixed-size and holistic representations, which can lead to degraded memory when modeling long sequences and limit precise information retrieval.  In contrast, self-attention retains full-token access and thus do not suffer from the two problems, but at the cost of heavy computation. This motivates us to explore an intermediate design that partially compresses the sequence while still maintaining global access.

% and use sparse attention or recurrence to capture long-range information. 

% Another line of work investigates the duality between linear RNNs and attention~\citep{linear_attention,mamba2,retnet,gatedlinearattention}, by enforcing strict linearity constraints and introducing the notion of state expansion, which represents recurrent states as matrices rather than vectors, and shifting from per-dimension to per-head gating.

% In contrast to them, we propose an alternative architecture. We argue that overusing attention in short-range contexts underutilizes its strengths and that such local processing can instead be handled more efficiently by lightweight structures such as recurrences. Based on this insight, we explore a design that lies between RNNs and attention.

% This design also avoids strict linearity assumptions on either side and adopts the classical recurrence form with per-dimension gating and no state expansion.

\begin{figure}[t]
\centering
    \begin{subfigure}[b]{0.73\linewidth}
    \centering
    \includegraphics[width=\textwidth]{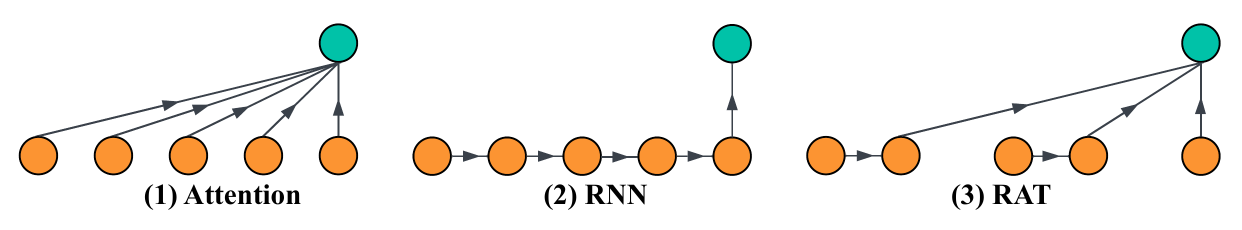}
    \caption{\textbf{Comparison of three structures}: Circles denote token representations and arrows show the information flow for the 5\textsuperscript{th} token. (1) Attention: Each token keeps its own representation and has full access to all previous tokens. (2) RNN: Information is progressively compressed along the sequence, and the current token accesses only the last hidden state. (3) \ratl{2}: Intra-chunk recurrence compresses local information, while inter-chunk attention enables direct access to previous chunks. (1) and (2) can be viewed as \rat with $L=1$ and $L=T$, respectively.
    }
    \label{fig:method}
  \end{subfigure}
  \hfill
  \begin{subfigure}[b]{0.25\textwidth}
    \centering
    \includegraphics[width=\textwidth]{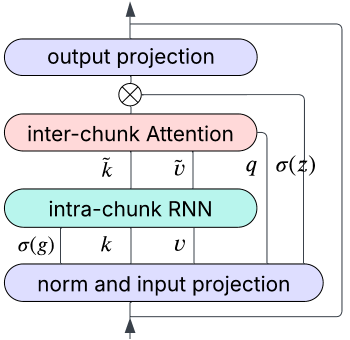}
    \caption{\textbf{\rat structure.} The symbol \(\sigma(\cdot)\) further applies a sigmoid function to the gates. 
    % Note that we do not use extra convolution or normalization in our design.
    }
    \label{fig:structure}
  \end{subfigure}
\end{figure}

We propose a \rat layer, a simple yet effective temporal mixing method with a chunk-based design. It divides long sequences into chunks, applying recurrence within each chunk for local modeling and softmax-based attention across chunks for direct access to distant information (see \autoref{fig:method}). Recurrence efficiently captures short-range dependencies while avoiding the memory degradation common in long sequences, whereas attention over chunk-level representations enables long-range retrieval. By adjusting the chunk size \(L\), \rat interpolates between attention (when \(L=1\)) and RNN (when \(L=T\)).

To make \rat scalable and efficient, we further explore its positional encoding, parameter allocation, and efficient implementations that specifically address the causal masking problem in this chunk-based design. Moreover, we investigate a hybrid architecture that interleaves \rat with sliding-window attention~\citep{longformer,gau,transformer-xl,block-recurrent}, which focuses computation on local windows. As \rat suggests that overusing attention in local contexts underutilizes its strengths, and such dependencies can be handled more efficiently via lightweight recurrence, the two approaches can be complementary.

We demonstrate that \rat is both efficient and performant. For instance, with a chunk size of \(L=16\), the latency of a single temporal mixing block at position 4096 is up to \(9\times\) lower than that of attention. As shown in \autoref{tab:main}, the full model achieves up to \(10\times\) higher maximum throughput, and interleaving with local attention yields about \(4\times\) improvement. We pretrain models at 1.3B scale and compare their results on 1) zero-shot commonsense reasoning with short contexts, (2) long-context understanding on LongBench~\citep{longbench}, and (3) supervised fine-tuning on long-context tasks. \rat with \(L=16\) performs on par with full attention in most benchmarks and even outperforms it on several LongBench tasks. When interleaved with local attention, it achieves the best overall results across all variants while maintaining high efficiency.
% We pretrained models with different chunk sizes at the 1.3B scale using a 4K context length, and observed that their perplexity~(PPL) falls between that of RNNs and attention. Then, we evaluated the models under three settings: (1) zero-shot commonsense reasoning tasks with short contexts, (2) LongBench tasks for long-context understanding, and (3) supervised fine-tuning on two question-answering and summarization tasks. The results show that \rat with \(L=16\) performs comparably to attention on most benchmarks, and even surpasses it on some LongBench tasks. When combined with local attention, the model achieves the strongest overall results among all variants, consistently improving most tasks while maintaining high efficiency.
Our main contributions are as follows:
\begin{enumerate}[nosep, leftmargin=*]
\item We propose the \rat layer, a novel intermediate architecture that bridges the efficiency of recurrence and the capacity of attention. It compresses only local context while preserving global access, enabling direct retrieval and avoiding the memory degradation caused by full sequence compression in long contexts.
  
\item \rat is simple, scalable, and efficient. It requires no custom CUDA kernels and is naturally compatible with existing multi-dimension parallelism schemes. We also introduce a hybrid variant by interleaving \rat with sliding-window attention, enabling efficient long-range modeling with strong local interactions.

\item We validate \rat through extensive experiments at the 1.3B scale across diverse tasks, including 7 short-context reasoning benchmarks, 11 long-context tasks, 4 supervised fine-tuning objectives, and 9 retrieval-heavy synthetic evaluations. \rat with \(L{=}16\) shows comparable performance to full attention with \(9\times\) faster single-layer decoding. Its hybrid variant with local attention yields the best overall results, along with a \(3-4\times\) maximum throughput boost—for instance, +1 accuracy on commonsense reasoning, +4 on code completion, +4 on a challenging QA task, and +1 on a difficult summarization task.
\end{enumerate}
\begin{table}[H]
\caption{Representative results for 1.3B models across pretraining, direct evaluation, and SFT. -SWA denotes interleaving with sliding-window attention~(SWA) (window size 1024). Maximum throughput is measured by generating 1024 tokens given a prompt of 3072 tokens on a H100 GPU in GH200 system. See \autoref{sec:exp} for details.
% Representative results for different architectures selected from pre-training, direct evaluation, and SFT of the 1.3B model. -SWA refers to the architecture in which certain layers are interleaved with the attention of sliding windows with a window size of 1024~\citep{longformer} every two layers.
% Val. denotes pre-training perplexity on a held-out validation set.
% CSR refers to six commonsense reasoning tasks.
% NQA\(^1\) indicates NarrativeQA (summary only).
% SQA, Summ, and Code represent single-document QA, summarization, and code completion tasks in LongBench~\citep{longbench}.
% Maximum throughput is measured by generating 1024 tokens given a prefill of 3072 tokens on a H100 GPU in GH200 system.
}
    \centering
    \vspace{-2mm}
    \begin{adjustbox}{max width=\textwidth}
    \begin{tabular}{l Hc cccc  cccc ccc}
    \toprule
    \textbf{Model} & \textbf{FLOPs} & \textbf{Throughput} 
    & \textbf{Pretrain} 
    & \multicolumn{4}{c}{\textbf{Direct Evaluation}} 
    & \multicolumn{3}{c}{\textbf{SFT}} \\
    \cmidrule(lr){4-4} \cmidrule(lr){5-8} \cmidrule(lr){9-10}
    & & & Val. & CSR & SQA & Summ & Code & NQA\(^1\) & QMSum \\
    & & token/sec & PPL & Avg. acc & Avg. F1 & Avg. Rouge-L & Avg. EditSum & F1 & Rouge-L  \\
    \midrule
    % RNN &  & 130K & 8.05 & 54.6 & 13.3 & 16.7 & 15.8 & 30.5 & 23.2 \\
    Attention & & 3052 & 7.61 & 56.9 & 18.2 & \underline{19.5} & \underline{23.9} & 61.3 & \underline{23.4}\\
    \ratrowall \ratl{16} &  & 31170 &7.67 & 56.7 & \bf 19.6 & \bf 20.2 & 17.4 & 60.8 & 23.3 \\
    \midrule
    Attention-SWA & & 4605 & \underline{7.61} & \underline{57.1} & 17.4 & 19.4 & 21.7 & \bf 63.3 & \underline{23.4}\\
    \ratrowall \ratlswa{16} & & 13582 & \bf 7.57 & \bf 58.0 & \underline{18.8} & \underline{19.5} & \bf 28.2 & \underline{63.2} & \bf 24.6\\
\bottomrule
\end{tabular}
\end{adjustbox}
\label{tab:main}
\end{table}

% Plan:
% \begin{itemize}
%     \item Language models have been dominated by Transformer architectures. However, recently, Mamba-style linear RNNs/SSMs attempted to dethrone them, and they have achieved impressive results (cite Jamba and Mamba)
%     \item Although Mamba-style approaches are more efficient at inference time, they are not as accurate as transformers, especially on retrieval-style tasks \cite{}. 
%     \item Here, we propose a new model called \rat{} which does ... Motivation for \rat{} is ...
%     \item Empirically \rat{} works really well. Provide results ...
%     \item Our contributions are: ...
% \end{itemize}

\section{Overall architecture}
To motivate our design, we first examine how attention and recurrence compress contextual information at the token level, along with the equations that will be reused later. We then introduce \rat as an intermediate mechanism that inherits the respective advantages of both.
\subsection{Attention: Full-Token Access}
In softmax attention mechanisms, each token has access to all preceding tokens through a learned token-specific weighted aggregation:
\begin{equation}
\label{eq:attention}
\begin{aligned}
\vy_t = f(\vq_t \mK^{\top}_{:}) \mV_{:}
\end{aligned}
\end{equation}
\( f(\cdot) \) denotes causal masking and the softmax function, and \(f(\vq_t \mK^{\top}_{:})\) represents the attention weights used to aggregate \(\mV_{:}\). We adopt Python-style indexing in \( \mK^{\top}_{:}, \mV_{:} \) to emphasize that each query attends to all keys and values. This full-token access makes \attn a dominant architecture in sequence modeling, but it suffers from high computational cost in both training and inference.
% Although \attn is widely considered a crucial component in state-of-the-art architectures for sequence modeling, its full-token access incurs a computational complexity of \( \mathcal{O}(T^2) \), where \( T \) is the length of the sequence. This quadratic growth in FLOPs can lead to substantial inefficiency during training and inference. We view attention as a \textbf{slow but lossless} module in the mixing of information between tokens and aims to organize it hierarchically with a \textbf{fast but lossy} component.

\subsection{Recurrence: Full-Sequence Compression}
Recurrent models maintain a summary of past information in a fixed-size representation, allowing each step to depend only on the previous state and the current input~\citep{gru,lstm}. To initiate the design of \rat, we adopt a simple and fast linear recurrence~\citep{minrnn,simple_linear_rnn,s5}, which uses diagonal matrices and no nonlinearities to simplify the classic recurrence into an EMA-like gating mechanism:
% RNNs exploit sequential dependencies in the data~\citep{gru,lstm}. As the model processes the sequence, it maintains a summary of past information in its hidden state, allowing each step to be computed solely from the previous state and the current input. This enables efficient inference and low FLOPs, in contrast to the full-sequence access required by \attn. To further improve training efficiency, recent work has explored linear RNN variants~\citep{mamba,lru}. For simplicity, we adopt the design proposed in~\citet{minrnn}, which is similar to that in~\citet{s5}. Using diagonal matrices %in place of full matrix multiplications 
% and removing nonlinearities, the recurrence reduces to an EMA-like gating behavior, which can be efficiently implemented with the parallel scan algorithm~\citep{s5}:
\begin{equation}
\label{eq:rnn}
\begin{aligned}
\tilde{\vv}_t &= \vg_t \odot \tilde{\vv}_{t-1} + (1 - \vg_t) \odot \vv_t, \\
\vy_t &= \vz_t \odot \tilde{\vv}_t,
\end{aligned}
\end{equation}
where \( \vg_t \) and \( \vz_t \in \mathbb{R}^{D} \) denote the per-dimension forget and output gates, respectively, computed via linear projections of the input followed by a sigmoid activation. Here, \(D\) refers to the dimension of the model. This can be efficiently implemented using the parallel scan algorithm~\citep{s5}. We take this minimal choice to highlight the core idea, but the recurrence in \rat is not limited to this certain form and can be extended to more advanced variants such as 2D recurrence or nonlinear RNNs. We leave this as future work. 

The state-based formulation of recurrence yields low computational cost and efficient inference, and performs well on short sequences. However, compressing the entire sequence history into a fixed-size and holistic state, even with more expressive designs, can still suffer from memory degradation when the sequence length grows~\citep{rnn_difficulty}, and limits precise information retrieval, particularly in noisy contexts.
% Thus, in this paper, we position \rnn as a \textbf{fast yet lossy} token mixing module.

%\subsection{Overview}
\subsection{\rat: Chunk-Based Intermediate Design }
To bridge the strong performance of attention enabled by full-token access with the efficiency of RNNs derived from full-sequence compression, we propose an intermediate design by reinterpreting the input as a sequence of shorter chunks. A recurrent module is applied within each chunk to model local dependencies, followed by cross-chunk attention to enable global interactions, as illustrated in \autoref{fig:method}. This design mitigates the fixed-size representation limitation of RNNs and the inefficiency of attention.

Technically, we divide a sequence of length \(T\) into \(C\) chunks of length \(L\), such that \(T = C \cdot L\). A token originally at position \( t \) is re-indexed as \( (c, l) \), where \( c \) denotes the chunk index and \( l \) the position within the chunk. Within each chunk, a forget gate \( \vg_{c,l} \) is used to recurrently aggregate the value \( \vv_{c,l} \) and key \( \vk_{c,l} \) vectors, yielding updated representations \( \tilde{\vv}_{c,l} \) and \( \tilde{\vk}_{c,l} \). For each query \( \vq_{c,l} \), we compute attention on the chunk-level key and value vectors, including \( \tilde{\mK}_{:,-1} \) for all preceding chunks, and \( \tilde{\vk}_{c,l} \) for the current chunk. The causal masking function \( f(\cdot) \) restricts attention to the chunks before it, followed by a softmax operation. Finally, an output gate is applied to produce the output:% following standard practice in gated RNNs.
\begin{equation}
\label{eq:rat}
\begin{aligned}
& \tilde{\vv}_{c,l}=\vg_{c,l}\odot \tilde{\vv}_{c,l-1} + (1 - \vg_{c,l})\odot \vv_{c,l} & \quad \text{(Intra-chunk \rnn)}\\
& \tilde{\vk}_{c,l}=\vg_{c,l}\odot \tilde{\vk}_{c,l-1} + (1 - \vg_{c,l})\odot \vk_{c,l} & \quad \text{(Intra-chunk \rnn)} \\
&\vy_{c,l}=f([\vq_{c,l}\tilde{\mK}_{:,-1}^{\top};\vq_{c,l}\tilde{\vk}_{c,l}^{\top}])[\tilde{\mV}_{:,-1};\tilde{\vv}_{c,l}] & \quad \text{(Inter-chunk \attn)} \\
& \vy_{c,l}=\vz_{c,l} \odot \vy_{c,l} &
\end{aligned}
\end{equation}

The overall computation flow is illustrated in \autoref{fig:structure}. Thanks to the short length of each chunk, the recurrent module efficiently captures local dependencies without significant information loss. Cross-chunk attention then enables global access without compressing distant information. This design preserves the accuracy of softmax-based attention while reducing computation, as inter-chunk attention operates over shorter sequences with \(L\) as the FLOPs reduction ratio. We also experimented with a reversed variant by applying attention within chunks and recurrence across them, but found that standard \rat achieves better FLOPs utilization and overall performance (see \autoref{appendix:sec:accuracy}).

%This design bridges the efficiency of \rnn and the accuracy of \attn. For accuracy, it avoids compressing all information into a single fixed-size hidden state, as in \rnn; instead, information is distributed across multiple chunk-level representations, enabling direct retrieval via cross-chunk \attn. For efficiency, it eliminates the need for full-sequence attention, which is computationally expensive, memory-intensive, and often unnecessary for modeling short-range dependencies.

\paragraph{Benefits of the chunk design} We position this as an \textbf{intermediate architecture} between RNNs and attention, and refer to it as \rat. By adjusting the chunk size \(L\), \rat interpolates between their behaviors: when \(L = T\), it reduces to a pure RNN; when \(L = 1\), it resembles full attention. This allows us to pursue the trade-off between \attn and \rnn with a single-layer design, and offers greater flexibility for hybrid modeling by using different chunk sizes to emulate varying behaviors.  From a mechanistic perspective, unlike either classic or advanced recurrence (state space or linear attention models) that rely on fixed-size and holistic representations, our chunk-based structure enables memory capacity to scale with sequence length while maintaining a fixed FLOPs reduction ratio. Its partial compression with direct access to prior chunks ensures its superior performance on retrieval-heavy tasks. 

% We position this design as an \textbf{intermediate architecture} design between RNN and attention, and refer to it as \rat. By adjusting the size of the chunk \(L\), \rat interpolates between the RNN and the attention behaviors: when \(L = T\), the model reduces to an RNN, since the entire sequence is treated as a single chunk; when \(L = 1\), it resembles the attention, since each token resides in its chunk. This chunk-based design allows \rat to maintain growing memory capacity as sequence length increases, while preserving a fixed FLOPs reduction ratio. Its partial compression and direct access to previous chunks also make it well suited for retrieval-heavy tasks. Moreover, the recurrence here can be extended to more advanced variants in the future work such as state space models (2D recurrence) for long chunks and nonlinear RNNs for short chunks where the training efficiency is no longer an issue. 

% In principle, if the chunk size \(L\) is sufficiently small, even non-linear RNNs could be viable in terms of efficiency, which we leave for future work. 

% Interestingly, we also explore the removal of positional encodings entirely, relying only on the causal mask at the attention level. When the number of chunks is small (for example, \(C = 64\) in a 200M model), performance remains stable. This suggests that \rat can be more robust to the absence of positional encodings than \attn, as \(C \ll T\), and such encodings are often linked to poor length generalization. Further discussion is provided in the Appendix.

\section{Scalable and Efficient Modeling with \rat}

\subsection{Design details}
\paragraph{Parameter allocation} 
We aim to control the parameters of \rat at \(4D^2\), given \(D^2\) for the output projection, \(3D^2\) for the query, key, and value projections in \attn, and an additional \(2D^2\) for the two gates in \rnn. We explore lightweight alternatives by using low-rank projections for the gates \(\vg\) and \(\vz\), or by sharing query and key projections across heads. Empirical results in \autoref{subsec:ablation} show that sharing the query \(\vq\) and key \(\vk\) slightly outperforms using low-rank gates. Notably, this design does not collapse into single-head attention, since the forget gate operates at the per-dimension level and produces distinct gated keys \(\tilde{\vk}\) for the inter-chunk attention.

\paragraph{Positional encoding}
We examine how to encode positional information in \rat, due to the presence of cross-chunk attention. Motivated by the fact that \rnn captures positional information through its sequential structure, we find that applying positional encoding at the chunk level, rather than relying on original token positions, yields slightly better fidelity. This strategy also improves length generalization, as the number of positions requiring encoding (i.e., the number of chunks) is much smaller than the full sequence length. In our main experiments, we use RoPE~\citep{rope} based on chunk indices for inter-chunk attention. In the length generalization study (see \autoref{appendix:subsec:generalization}), we further explore NoPE~\citep{nope}, which yields the best overall generalization performance. 

% Both RoPE and NoPE are compatible with existing techniques for improving the length generalization of attention.

\paragraph{Hybrid design with Local Attention} 
\rat is a hierarchical architecture, but not a hybrid model, as it applies the same strategy across all tokens, layers, and heads. It is compatible with various hybrid strategies and offers more flexibilities in hybrid modeling by varying the chunk size in different layers or heads. In particular, we explore interleaving \rat with sliding-window attention (SWA)~\citep{longformer,gau,transformer-xl}, a widely adopted technique in recent models~\citep{cohere_llm,samba,gateddeltanet,griffin}. We find that the two are highly complementary: while local attention methods allocate most computation within fixed windows, \rat reserves attention for global access and handles local modeling more efficiently. Interleaving them enables the model to efficiently and effectively capture both short-range and long-range dependencies.

\subsection{Efficiency}
We discuss efficiency-related aspects of \rat, and show that our current implementation does not rely on custom CUDA or Triton kernels, yet achieves significant speed-ups in both training and generation in experiments. The pseudocode for training is provided in \autoref{algo:train_or_prefill}, and the decoding algorithm is shown in \autoref{algo:gen}.

To begin, the FLOPs per token of \rat are \(\mathcal{O}(C \cdot D)\), compared to \(\mathcal{O}(D)\) for \rnn and \(\mathcal{O}(T \cdot D)\) for full attention, where \(C\) is the number of chunks, \(D\) the model dimension, and \(T\) the sequence length. Most components of \rat are simple and easy to implement, except that the chunk-based design introduces a non-trivial causal masking challenge during training, which we address below.

\paragraph{Causal masking problem}
In training, where tokens are processed in parallel, special care must be taken to apply causal masking in inter-chunk \attn. First, a block-wise causal mask is required. Second, each token must also attend to its own chunk's key and value, which should be gated up to its position. They vary across tokens due to causal masking, preventing efficient parallel computation. To address this, we adopt an online softmax formulation~\citep{onlinesoftmax}: we separately compute \( f(\vq_{c,l} \tilde{\mK}_{:,-1}^\top) \tilde{\mV}_{:,-1} \) and \( f(\vq_{c,l} \tilde{\vk}_{c,l}^\top) \tilde{\vv}_{c,l} \), and then combine the results by adjusting the softmax denominator. The first term can be implemented in parallel using existing attention frameworks, while the second is handled via a simple einsum.

\paragraph{Practical implementation}
For training, we implement intra-chunk recurrence in \autoref{eq:rat} using PyTorch’s \textit{associative scan}, enabling forward and backward passes with \(\mathcal{O}(T)\) FLOPs. Compared to full-sequence recurrence, chunking reduces scan depth from \(\mathcal{O}(\log T)\) to \(\mathcal{O}(\log L)\) and thus improves parallelism. For inter-chunk \attn, we use PyTorch's \textit{flex attention} to implement the first term above. It supports \textit{flash attention} with flexible features such as custom masks and returning the softmax denominator, and thus aligns well with our needs. For decoding, tokens are generated sequentially. Intra-chunk recurrence only requires single-step updates and can be implemented directly. For inter-chunk \attn, standard implementations like \textit{flash attention}~\citep{flashattention} can be used without modification, as no complex causal masking is required at inference time. 

% In this setting, a sequence of tokens is processed in parallel, requiring the model to support efficient parallel computation. For the intra-chunk recurrent component in \autoref{eq:rat}, we use PyTorch's higher order operator \textit{ associative scan} to implement forward and backward passes with \(\mathcal{O}(T)\) FLOPs. For inter-chunk \attn, each token also attends to its own chunk’s key and value, which vary due to causal masking. To handle this efficiently, we adopt an online softmax formulation~\citep{onlinesoftmax}: we separately compute \( f(\vq_{c,l} \tilde{\mK}_{:,-1}^\top) \tilde{\mV}_{:,-1} \) and \( f(\vq_{c,l} \tilde{\vk}_{c,l}^\top) \tilde{\vv}_{c,l} \), then combine them by adjusting the softmax denominator. The first term is computed efficiently and in parallel using PyTorch’s \textit{flex attention}, which supports custom causal masks and returns the softmax normalization term. The second term is implemented via a simple einsum.

% \paragraph{Decoding efficiency}  
\paragraph{Parallelism}
We think \rat is compatible with both tensor parallelism and context parallelism, which are commonly used for large model dimensions and long sequence lengths, respectively. Both the recurrence and cross-chunk attention in \rat are head-independent, making it easy to apply standard tensor parallelism by assigning different heads to different GPUs. For context parallelism, the intra-chunk recurrence is chunk-independent, allowing chunks to be distributed across GPUs. Since \rat stores much fewer number of chunk-level key/value vectors (e.g., 16× fewer than full attention), the cross-chunk attention may even avoid ring-style communication.

\section{Experiments}
\label{sec:exp}
In this section, we present the efficiency and large-scale evaluations of \rat, along with comparisons to other models. Additional discussions are provided in \autoref{appendix:sec:accuracy}.

\subsection{Setup}
For brevity, we summarize the setup for the 1.3B model with a 4K context window, which is used in most of our experiments. Full implementation details are available in \autoref{appendix:sec:impl}.

\paragraph{Model}
We adopt a Transformer architecture that interleaves a token mixing block with a hidden-state mixing block (FFN), each wrapped with residual connections and LayerNorm. We compare variants that use different token mixing modules, including \rnn, \attn, and \rat, as well as recent state space and linear attention models. In addition, we provide hybrid models that interleave \rat with sliding-window attention (window size 1024), such as \attnswa and \ratswa.
% This substitution reduces both the FLOPs and the KV cache size per layer. 
% For model scales, we first use a 200M parameter model trained on the long-context PG19 book dataset~\citep{pg19} to explore architectural design and demonstrate the potential of \rat over \rnn. All other experiments are conducted with 1.3B-parameter models, which serve as the primary focus of this work.  

\paragraph{Efficiency}
We benchmark the latency of a single token mixing block, including input and output projections, on a single H100 GPU (GH200 system, 120GB), and also report the maximum throughput of the full model. We measure the time required to train on a full sequence or to generate a batch of tokens at specific positions. For fair comparison, we use \textit{flash attention} for the \attn baseline and \textit{associative scan} for \rnn. All models are compiled using \textit{torch.compile} and evaluated in bfloat16 with \textit{torch.cuda.amp}.

\paragraph{Accuracy}
We pretrain all 1.3B models on 100B tokens from FineWeb-Edu~\citep{fineweb_edu} using the same setup: a learning rate of 8.0e-4 decayed to 1.0e-6 (cosine schedule) and a global batch size of 2M tokens, following DeepSeek's hyperparameter guidelines~\citep{deepseek_hp}. We directly evaluate the models on both short-context tasks—several classical commonsense reasoning benchmarks~\citep{eval-harness}—and long-context tasks, including 11 tasks from LongBench~\citep{longbench} covering QA, summarization, and code completion. Since LongBench includes instruction-heavy prompts that pretrained models often struggle with, we also include SFT-based evaluations. Specifically, we use NarrativeQA~\citep{narrativeqa} (two modes), QMSum~\citep{qmsum}, and WikiSum~\citep{wikisum} to test long-context understanding with SFT. To assess retrieval capabilities, we include the Ruler benchmark~\citep{ruler} and test nine synthetic needle-in-haystack tasks with varying configurations. A single round of lightweight fine-tuning is applied to adapt models to specific prompts. 
% Models are trained in official splits with an answer-only loss and evaluated in test sets. 
% We found that different hyperparameter choices usually yield similar trends; thus, we fix the learning rate and batch size to (1.0e-5, 128) for large datasets and (1.0e-5, 32) for the smaller QMSum~\citep{qmsum}, following common practice for 1B-scale models. We sweep over the number of epochs and report the best result for each model. 
% In practice, QA tasks typically converge within 1–2 epochs, while summarization benefits from slightly longer training.

\subsection{Analyses of \rat}
\begin{minipage}[htbp]{1.0\textwidth}
  \begin{minipage}[t]{0.63\textwidth}
    \centering
    \captionof{figure}[t]{\textbf{Latency of the temporal mixing block} (including linear projections) with a model dimension of 2048. (a): full-sequence latency with 200K tokens; (b): generation of 512 tokens at specified positions. We adopt \textit{flash attention} for Attention.}
    \includegraphics[width=\textwidth]{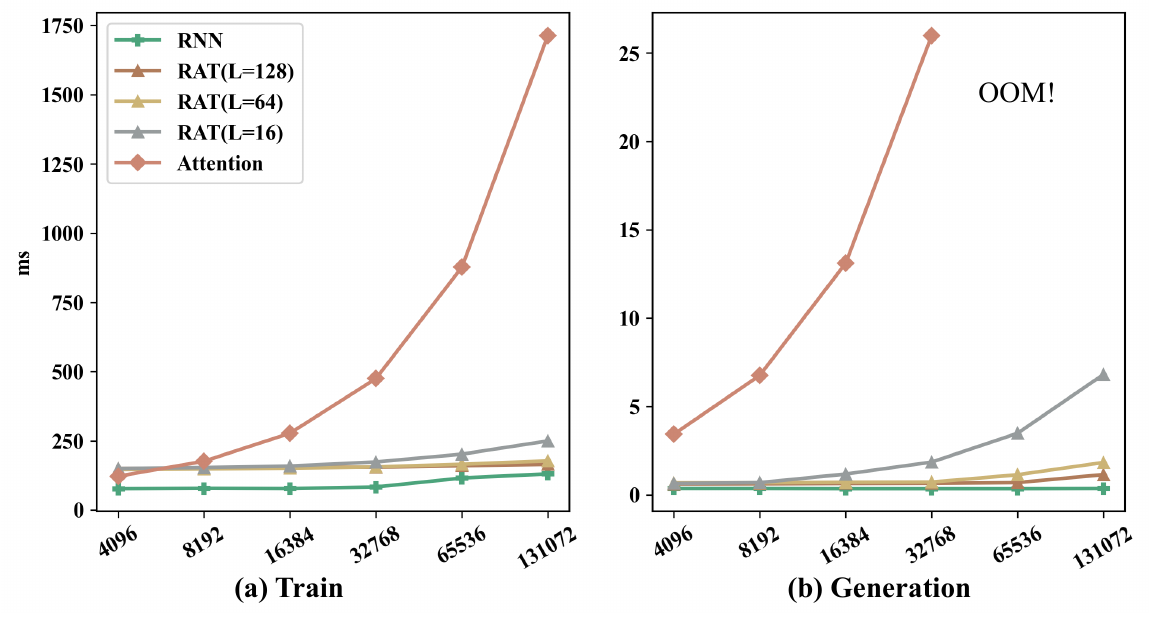}
    \label{fig:eff}
  \end{minipage}
  \hfill
  \begin{minipage}[t]{0.35\textwidth}
    \centering
    \captionof{table}[t]{\textbf{Maximum throughput of full models} (tokens/sec), measured by generating 1024 tokens from a 3072-token prompt. By reducing the KV cache memory and boosting speed, we achieve 10× maximum throughput compared to \textit{flash attention} , and even more on 13B models, as \attn suffers from poor GPU utilization at larger scale.}
    \label{tab:maximum_throughput}
    \begin{adjustbox}{max width=\textwidth}
      \begin{tabular}{lccc}
        \toprule
        \bf Model & \bf 1.3B & \bf 7B  & \bf 13B \\
        \midrule
        \ratl{16} & 31170  & 10103 & 5749 \\
        Attention & 3152 & 983 & 534  \\
        Ratio & 10.2\(\times\) & 10.3\(\times\) & 10.8\(\times\) \\
        \bottomrule
      \end{tabular}
    \end{adjustbox}
  \end{minipage}
\end{minipage}
\paragraph{Efficiency study}
\label{subsec:efficiency}
\autoref{fig:eff} and \autoref{tab:maximum_throughput} present the efficiency comparison, with additional results provided in \autoref{appendix:sec:efficiency}. For training, on the strong H100 GPU, when the sequence length is short (e.g., 4096), \rat is slightly slower than attention due to underutilized GPU parallelism in the \textit{flex attention} (with only 256 chunks) and the overhead introduced by \textit{associative scan}. We expect these can be improved through further kernel-level optimizations. As the sequence length increases, \rat becomes increasingly efficient: for \(L=16\), we observe approximately \(2\times\) speedup at 16K, \(3\times\) at 32K, \(4\times\) at 64K, and \(7\times\) at 100K tokens. For generation, \ratl{16} achieves a \(9\times\) speedup at position 4K and around \(10\times\) for longer sequences. It also reduces KV cache usage, making it significantly less prone to out-of-memory (OOM) errors and enabling much higher maximum throughput.

\paragraph{Ablation study}
\label{subsec:ablation}
We conduct the ablation study on a 200M-parameter model trained on the book dataset. Allocating more parameters to the intra-chunk \rnn gates yields significantly better performance than assigning them to the inter-chunk \attn query and key projections, improving the perplexity by 0.4–0.5. Further replacing the original RoPE with a cross-chunk variant, where positions are indexed by chunk index, brings additional improvement, especially at long sequence lengths, with a 0.3 drop in PPL observed.

\begin{figure}[htbp]
    \centering
    \includegraphics[width=\textwidth]{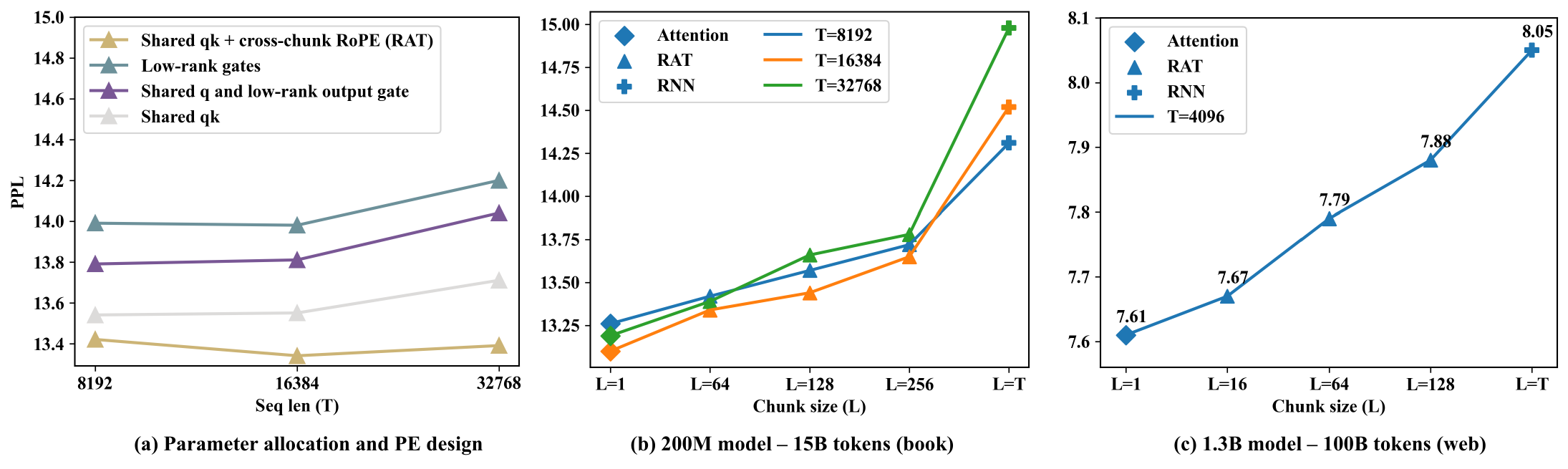}
     \caption{(a) Ablation study on \ratl{64}. (b) and (c) show pretraining results on 200M and 1.3B models, respectively. \rat lies between \rnn and attention in terms of pretraining perplexity.}
\label{fig:ablate_pretrain}
\end{figure}

\paragraph{Pretraining}
\label{subsec:accuracy}
In \autoref{fig:ablate_pretrain}(b) and (c), we begin our study by examining pretraining perplexity as a foundational measure of architectural performance, demonstrating that \rat falls between \attn and \rnn. In (b), using a 200M model, we observe that increasing the chunk size leads to higher perplexity while reducing FLOPs. Viewing \attn as \(L=1\) and \rnn as \(L=T\), we observe perplexities of 13.26 for \attn, 13.42 for \ratl{64}, 13.72 for \ratl{256}, and 14.31 for \rnn. Interestingly, increasing the training context from \(T = 8192\) to \(T = 32768\) leads to a sharp rise in perplexity for \rnn, while both \attn and \rat remain stable, demonstrating the pitfalls of full-sequence compression. We then scale up to a 1.3B model trained on web data, which primarily consists of short sequences (often under 1K tokens) concatenated with separator tokens to construct 4K-length contexts. We find that \ratl{16} gives the comparable performance to \attn, and again \rat with different chunk sizes exhibits behavior intermediate between \attn and \rnn.

\subsection{Large-scale evaluation and comparisons}
In this section, we present direct evaluation and SFT results of our trained 1.3B models. We mainly compare \ratl{16} with its two extremes: \rnn and \attn. Additionally, we include results from recent state space and linear attention models~\citep{gateddeltanet}, which are trained under the same settings (model size, sequence length, and dataset) as ours. We report the value/slot sizes, which indicates the size of memory slots and FLOPs, based on sequence length \(T\), model dimension \(D\), and number of layers \(N\). Note that state space and linear attention models use fixed-size memory and expand the classic recurrence state, while \rat scales its memory capacity with the sequence length.
\paragraph{Commonsense reasoning}
In \autoref{tab:csr_4k}, the performance gap~(within two points on average) between \rnn and \attn is small on these short commonsense reasoning tasks, indicating that \rnn can be surprisingly competitive in short-context settings. \ratl{16} consistently outperforms \rnn on most tasks and surpasses other recent models on 4 out of 7 benchmarks, with only at most \(20ND\) memory slots. We also observe that some binary-choice tasks (e.g., BoolQ, WinoGrande) show inconsistent results under small-scale pretraining. For instance, \attn struggles on BoolQ, while \rat underperforms on WinoGrande. Interestingly, the hybrid variant \ratlswa{16}, which interleaves \rat with \swa, achieves the best overall accuracy across almost all tasks. This suggests a complementary architecture: using \swa for strong local attention and \rat for long-range dependencies can bring us a model that is not only more efficient but also more accurate than its dense counterpart.
\begin{table}[htbp]
\centering
\caption{We evaluate our models with model dimension \(D=2048\) and number of layers \(N=24\) on seven commonsense reasoning tasks using \textit{lm-evaluation-harness}~\citep{eval-harness}. Since all tasks (except BoolQ, which has only 13 sequences \(>300\) tokens) have sequences \(\leq 300\) tokens, we set \(T=300\). Results marked with * are from a recent study~\citep{gateddeltanet} trained on the same setting as ours. For SWA, slot sizes remain \(150ND\) due to the short context.}
\label{tab:csr_4k}

\resizebox{1.0\textwidth}{!}{
\begin{tabular}{llccccccc}
\toprule
\textbf{Model} & \textbf{Value/Slot size}
& \textbf{ARC-C} & \textbf{ARC-E}  & \textbf{HellaSwag} & \textbf{LAMBADA}  & \textbf{PIQA} 
& \textbf{WinoGrande} & \bf BoolQ \\
 & \(T=300\) & acc/acc\_n & acc/acc\_n & acc/acc\_n & acc & acc/acc\_n & acc & acc\\
\midrule
\rnn & \(ND\) & 35.15/38.57 & 70.66/66.46 & 42.43/55.95 & 39.74 & 71.49/72.69 & 53.99 & 62.54\\
Mamba* & \(64ND\) & -/35.40 & 69.52/- & -/52.91 & -/43.98 & 71.32/- & 52.95 & 61.13\\ 
Mamba2* & \(256ND\) & -/37.88 & 72.47/- & -/55.67 & -/45.66 & 71.87/- & 55.24 & 60.13\\
DeltaNet* & \(128ND\) & -/35.66 & 68.47/- & -/50.93 & -/42.46 & 70.72/- & 53.35 & 55.29\\
GatedDeltaNet* & \(288ND\) & -/38.39 & 71.21/- & -/55.76 & -/46.65 & 72.25/- & 57.45 & 60.24\\
Attention & \(300ND\) & 35.67/37.71 & 71.25/66.67 & \bf 44.16/57.44 & \bf 47.84 & \bf 73.29/72.91 & \bf 57.70 & 58.23\\
\ratrowall \ratl{16} & \(\sim20ND\) & \bf 36.35/39.33 & \bf 72.64/67.80 & 43.27/56.44 & 44.40 & 72.03/72.69 & 53.20 & \bf 63.30\\
\midrule
GatedDeltaNet-SWA* & \(288N/2D-300N/2D\) & -/40.10 & 71.75/- & -/56.53 & 47.73 & 72.57/- & \bf 58.4 & 63.21\\
Attention-SWA & \(300N/2D-300N/2D\) & 35.24/37.20 & 71.97/66.20 & 44.36/57.09 & 47.97 & 72.25/72.74 & 56.91 & 61.25 \\
\ratrowall \ratlswa{16} & \(\sim 20N/2D-300N/2D\) & \bf 37.20/40.78 & \bf 72.56/68.22 & \bf 44.33/57.85 & \bf 49.33 & \bf 73.29/73.94 & 56.91 & \bf 63.21\\
\bottomrule
\end{tabular}
}
\end{table}

\begin{table}[htbp]
\centering
\caption{Evaluation results on LongBench with \(T=4096\), \(D=2048\), and \(N=24\). NQA: NarrativeQA, MQA: MultiFieldQA-en, HQA: HotpotQA, WQA: 2WikiMultihopQA, MSQ: Musique, GR: GovReport, MN: MultiNews, TQA: TriviaQA, RBP: RepoBench-P. Metrics: F1 for QA tasks, Rouge-L for summarization, and EditSum for code tasks. Results marked with * are from a recent study~\citep{gateddeltanet} trained under the same settings as ours. Note that they use local attention with a window size of 2048, while we use 1024.
}
% }
\resizebox{\textwidth}{!}{%
\begin{tabular}{l l cccc cccc cccc ccc}
\toprule
\multirow{3}{*}{\bf {Model}}
& \bf Value/Slot size & \multicolumn{4}{c}{\textbf{Single-Document QA}} 
& \multicolumn{4}{c}{\textbf{Multi-Document QA}} 
& \multicolumn{4}{c}{\textbf{Summarization}} 
& \multicolumn{3}{c}{\textbf{Code Completion}} \\
\cmidrule(lr){3-6} \cmidrule(lr){7-10} \cmidrule(lr){11-14} \cmidrule(lr){15-17}
& \(T=4096\) & NQA & Qasper & MQA & Avg.
& HQA & WQA & MSQ & Avg. 
& GR & QMSum & MN & Avg. 
& LCC & RBP & Avg.\\
% & \bf NQA & \bf Qasper & \bf MQA & \bf Avg.
% & \bf HQA & \bf WQA & \bf MSQ & \bf Avg.
% & \bf GR & \bf QMSum  & \bf \bf MN & \bf Avg.
% & \bf TREC & \bf TQA & \bf SAMSum & \bf Avg.
% & \bf LCC & \bf RBP & \bf Avg. \\
% & F1 & F1 & F1 & -
% & F1 & F1 & F1 & -
% & Rouge-L & Rouge-L  & Rouge-L& -
% & Acc & F1 & Rouge-L & -
% & EditSum & EditSum & -\\

\midrule
\rnn & \(ND\) & 11.7 & 8.8 & 19.5 & 13.3 & 9.1 & 15.1 & 5.4 & 9.9 & 16.9 & 16.8 & 16.4 & 16.7 & 14.0 & 17.6 & 15.8 \\
Mamba2* & \(256ND\) & 11.1 & 11.3 & 18.6 & 13.7 & 11.8 & 15.1 & \bf 6.7 & 11.2 &  6.7 & 14.5 & 7.4 & 9.5 & 17.9 & 20.6 & 19.3 \\
DeltaNet* & \(128ND\) & 12.9 & 10.8 & 21.5 & 15.1 & 10.9 & 13.2 & 5.1 & 9.7 & 6.5 & 13.5 & 7.2 & 9.1 & 17.6 & 20.3 & 19.0 \\
GatedDeltaNet* & \(288ND\) & 14.1 & 14.0 & 23.3 & 17.1 & 13.7 & 14.4 & 5.8 & 11.3 & 7.5	& 16.4 & 7.9 & 10.6 & 18.7	& 22.1 & 20.4 \\
Attention & \(4096ND\) & 12.3 & 14.0 & 28.2 & 18.2 & 13.4 & 17.7 & 4.9 & 12.0 & 18.6 & \bf 18.9 & \bf 21.0 & 19.5 & \bf 21.3 & \bf 26.5 & \bf 23.9 \\
\ratl{64} & \(64ND\) & 13.9 & 13.0	& 27.0 & 18.0 & 14.9 & 16.3 & 4.3 & 11.8 & 18.9 & 17.7 & 19.9	& 18.8 &  16.2 & 19.7 & 18.0 \\
\ratrowall \ratl{16} & \(256ND\) & \bf 14.5 & \bf 16.1 & \bf 28.3 & \bf 19.6 & \bf 16.7 & \bf 18.9 & 6.3 & \bf 14.0 & \bf 24.9 & 17.5 & 18.3 & \bf 20.2 & 14.2 & 20.6 & 17.4 \\
\midrule
GatedDeltaNet-SWA* & \(288N/2D-2048N/2D\) & \bf 14.5 & 12.3 & 26.6 & 17.8 & 12.6 & \bf 23.6 & 6.1 & \bf 14.1 & 9.1 & 16.1 & 12.8 & 12.7 & 15.5 & 19.2 & 17.4 \\
\attnswa & \(4096N/2D-1024N/2D\) & 13.1 & 14.6 & 24.5 & 17.4 & 13.7 & 19.0 & 6.1 & 12.9 & 19.4 & 17.3 & \bf 21.4 & 19.4 & 18.7 & 24.6 & 21.7 \\
\ratrowall \ratlswa{16} & \(256N/2D-1024N/2D\) & 12.7 & \bf 15.6 & \bf 28.2 & \bf 18.8 & \bf 14.2 & 18.1 & \bf 7.4 & 13.2 & \bf 20.1 & \bf 18.6 & 19.8 & \bf 19.5 & \bf 26.3 & \bf 30.1 & \bf 28.2 \\
\bottomrule
\label{tab:longbench_4k}
\end{tabular}%
}
\end{table}

\paragraph{LongBench}
In \autoref{tab:longbench_4k}, we present results on 11 long-context tasks from LongBench. Since LongBench includes instruction-heavy prompts across diverse domains, it's natural that no single model dominates all tasks. Evaluating purely pretrained models on such instruction-based datasets can be very challenging. Therefore, we focus on general performance trends, especially average scores by task category. As shown in the table, \ratl{16} and its hybrid variant \ratlswa{16} achieve top performance in many groups, such as question answering and summarization. This is noteworthy, given that \ratl{16} slightly lags behind \attn in pretraining perplexity and underperforms in short-context commonsense reasoning. We hypothesize that its advantage on LongBench stems from structural adaptation to very long sequences—for instance, achieving 24.9 on GovReport~(3-1) (vs. 18.6 for the \attn baseline) and 16.7 on HotpotQA~(2-1) (vs. 13.4). Second, unlike the small gap (\(\sim\)2 points) between \rnn and \attn on short-context tasks, the difference becomes much larger on LongBench, further demonstrating the problem of full-sequence compression. We also notice that \ratl{16} performs less favorably on code completion. However, its hybrid variant significantly improves performance in this setting, outperforming the second-best model by about four points.

\paragraph{Supervised Fine-tuning}
Since LongBench tasks are instruction-based and pretrained-only models often struggle to follow specific prompts, we further assess the performance of SFT on some long-context QA and summarization tasks, as shown in \autoref{tab:sft_4k}. For NarrativeQA, where passages are extremely long and may not contain the answer within the truncated 4K-token context, we evaluate two settings: (1) summary-only, which uses only the passage summary, and (2) summary plus passage, which concatenates the summary with the full passage. In both cases, \ratlswa{16} delivers strong performance while remaining efficient. Although both the summary and passage contain useful information, we observe that models, particularly the attention-based ones, perform worse in the second setting, with F1 dropping to 33. We hypothesize that in this setup, \attn may be more easily distracted due to the absence of structural bias, which can lead to flattened attention score distributions~\citep{scalablesoftmax}. In contrast, models based on \rat appear more robust, likely due to their chunk-wise structure. For summarization tasks, \rnn performs better than in QA, and \ratlswa{16} achieves the best result on QMSum, surpassing the next-best model by approximately 1 Rouge-L point.

\begin{table}[htbp]
\caption{\textbf{Left}: Average and 95th percentile of context lengths  
of datasets with the LLaMA2 tokenizer.  
\textbf{Right}: SFT performance on long-context datasets with \(T=4096\), \(D=2048\), and \(N=24\).  
F1 is used for QA tasks, and Rouge-L is used for summarization tasks.  
NarrativeQA$^{\textbf{1}}$: only the summary of a passage is provided.  
NarrativeQA$^{\textbf{2}}$: the summary plus the full passage is provided; the passage typically contains much irrelevant information.}

\begin{minipage}[t]{0.32\textwidth}
\centering
\label{tab:sft_token_4k}
\begin{adjustbox}{max width=\textwidth}
\begin{tabular}{lcc}
\toprule
\bf Task & \bf Avg. & \bf 95th pctl. \\
\midrule
NarrativeQA\(^{\textbf{1}}\) & 838 & 1296 \\
NarrativeQA\(^{\textbf{2}}\) &  97k & 254k\\
QMSum & 16k & 29k \\ 
WikiSum & 1828 & 3473\\
\bottomrule
\end{tabular}
\end{adjustbox}
\end{minipage}
\hfill
\begin{minipage}[t]{0.66\textwidth}
\centering
\label{tab:sft_4k}
\begin{adjustbox}{max width=\textwidth}
\begin{tabular}{llcccc}
\toprule
\bf Model & \bf Value/Slot size & \bf NarrativeQA\(^{\textbf{1}}\) & \bf NarrativeQA\(^{\textbf{2}}\) & \bf QMSum & \bf WikiSum \\
% & F1 & F1 & Rouge-L & Rouge-L \\
\midrule
\rnn & \(ND\)  &30.5 & 27.4 & 23.2 & 34.3 \\
\ratrowall \ratl{16} & \(256ND\) & 60.8 & \underline{43.5} & 23.3 & 36.6 \\
Attention & \(4096ND\) & 61.3 & 33.0 & \underline{23.4} & \bf 36.8 \\
\midrule
Attention-SWA & \(4096N/2D-1024N/2D\) & \bf 63.3 & 39.6 & \underline{23.4} & \underline{36.7} \\
\ratrowall \ratlswa{16} & \(256N/2D-1024N/2D\) & \underline{63.2} & \bf 43.7 & \bf 24.6 & \underline{36.7} \\
\bottomrule
\end{tabular}
\end{adjustbox}
\end{minipage}
\end{table}
\begin{table}[htbp!]
\centering
\caption{\textbf{Retrieval ability}: Accuracy performance with exact match scoring on the Needle-in-Haystack tasks with different configurations from the \textsc{RULER} benchmark~\citep{ruler}. We use \(T=4096\), \(D=2048\), and \(N=24\). }
\label{tab:niah}
\resizebox{\textwidth}{!}{
\begin{tabular}{ll ccc ccc cc}
\toprule
\bf Model & \bf Value/Slot size & \bf \texttt{single\_1} & \bf \texttt{single\_2} & \bf \texttt{single\_3} & \bf \texttt{multikey\_1} & \bf \texttt{multikey\_2} & \bf \texttt{multikey\_3} & \bf \texttt{multiquery} & \bf \texttt{multivalue} \\
\midrule
\rnn & \(ND\) & 99.4 & 99.8 & 0.0 & 2.2 & 0.0 & 0.0 & 5.8 & 5.7\\
\ratl{64} & \(64ND\) & 100.0 & 99.8 & 55.6 & 96.4 & 62.2 & 1.0 & 86.3 & 89.1\\
\ratl{16} & \(256ND\) & 99.6 & 100.0 & 94.6 & 99.6 & 99.6 & 82.6 & 91.2 & 94.8\\
Attention & \(4096ND\) & 100.0 & 100.0 & 100.0 & 99.6 & 99.6 & 95.2 & 98.6 & 99.1 \\
\bottomrule
\end{tabular}
}
\end{table}

\paragraph{Retrieval ability}
In \autoref{tab:niah}, we evaluate the three models on challenging synthetic needle-in-haystack tasks that test retrieval ability (see \autoref{appendix:subsec:evaluation} for task descriptions). It can be seen that \attn consistently performs well due to its full-token access. \rnn handles simpler tasks like \texttt{single\_1} and \texttt{single\_2} reasonably well, but its performance drops on harder settings. \ratl{16} matches attention closely on most tasks but struggles with UUID ones (\texttt{single\_3}, \texttt{multikey\_3}) due to their difficulty and evaluation metric as exact match scoring. Meanwhile, \ratl{64} falls between RNNs and \ratl{16}, as expected given its partial access to long-range context. These results align with the model structures: attention sees all tokens directly, recurrence compresses all history into a single state, and \rat blends both, preserving chunk-level access while reducing FLOPs.

% state space and linear attention models
  % state space
  % linear attention
  % log-style
  % hybrid models

% Softmax-based attention
  % local attention, GAU and chunking works
  % GSA
\section{Related work}
\label{sec:related_work}
\paragraph{State space and linear attention models}
Recurrent neural networks (RNNs) have long been used for sequential modeling~\citep{elman1990finding, siegelmann1992computational, bengio1994learning}. Recent state space models~\citep{s4, s5, mamba, griffin, lru} extend recurrence by expanding hidden states from vectors to matrices for higher accuracy, removing non-linearity and using structured operations (e.g., diagonal matrices) to enable parallel computation~\citep{s5}. Per-head gating connects these models~\citep{mamba2} to linear attention~\citep{linear_attention,retnet,rwkv,lru,deltanet,gateddeltanet} and shows the duality between linear recurrence and linear attention. Unlike this duality perspective, we take an intermediate view between per-dimension-gated RNNs and softmax-based attention, without enforcing strict linearity. Linear recurrence in \rat is intended primarily for training efficiency, rather than imposed as a design constraint.

Despite state expansion, these models inherit the fixed-size representation limitation of classic RNNs, leading to degraded memory on long sequences. A concurrent work~\citep{log_linear_attention} proposes addressing this by growing the memory slots logarithmically. We similarly target the fixed-size memory problem, but use softmax-based attention to access distant tokens. Hybrid strategies~\citep{samba,hymda} combining state space models and attention are also popular. While \rat is not a hybrid with the same computation graph for all tokens and all layers, it is orthogonal to such designs and provides more flexibility, such as using different chunk sizes for different layers and heads, which we leave for future work.

\paragraph{Softmax-based attention}
Attention mechanisms suffer from slow computation due to their full-token access. Earlier works have explored local attention with chunking, combined with recurrence or sparse mechanisms to access earlier context~\citep{longformer,gau,mega,megalodon,landmarkattention}. For example, \citet{mega} gates query/key vectors before local attention, while \citet{gau} adds the outputs of local attention to cross-chunk linear attention. In our work, we instead leverage attention's strength in directly accessing distant tokens with lightweight recurrence for local contexts, and hierarchically organize their outputs. As shown in \autoref{appendix:subsec:ablation}, \rat utilizes FLOPs more effectively than local attention. We also regard \rat and local attention as structurally complementary. Although we experimented with interleaving them in this work, we leave the combination with more advanced local attention variants in future research.

\rat can be viewed as an attention mechanism with resolution, akin to how humans process information: short-term inputs are integrated into coherent representations, while long-term events are stored as memory anchors for selective retrieval. Compared to dilated attention~\citep{longformer}, our simple recurrence enables global perception, and the chunk-level \attn has high training parallelism and reduced cache memory (see \autoref{appendix:subsec:ablation} for more). Finally, a recent work~\citep{gatedslotattention} also aggregates the key and value vectors and applies softmax-based attention over them. However, it compresses the full-sequence vectors into a 2D memory matrix, unlike our chunk-based design, which enables flexible memory slots.

\section{Conclusion}
\label{sec:conclusion}
This paper proposes the \rat structure, an intermediate architecture between RNN and attention. It segments sequences into chunks, applies \rnn within each chunk to capture local dependencies, and employs \attn across chunks to model long-range dependencies. We detail its architectural design and investigate its combination with the sliding-window attention mechanism. Experimental results across various settings demonstrate the efficiency and accuracy benefits of \rat, highlighting its potential for future language model development.

\textbf{Limitations.} 
Due to resource constraints and the exploratory nature of this work, our experiments are limited to models with up to 1.3B parameters. We have not yet scaled \rat to larger language models such as 7B or 14B to confirm our results hold at that scale.  
Additionally, we did not adopt supervised fine-tuning techniques commonly used in current industry practices, which typically involve substantial data resources and many engineering details. Instead, we followed a more classical approach based on train-test splits to evaluate performance, as our focus lies primarily on architectural design during pretraining.  
Lastly, \rat may still face length generalization issues from positional encoding, similar to attention. But as shown in \autoref{appendix:subsec:generalization}, this can be significantly reduced by shorter inter-chunk attention spans, and we also explore the use of NoPE.
% Lastly, the \rat layer may still encounter generalization challenges similar to those in attention due to the use of positional encoding. However, as discussed in \autoref{appendix:subsec:generalization}, the issue appears significantly reduced due to the shortened attention span across chunks, and we also explore the use of NoPE.
% \section*{References}
\begin{ack}
We sincerely thank the RCP and the SCITAS team at EPFL for GPU support. We also thank the Swiss AI Initiative and the Swiss National Supercomputing Centre (CSCS) for supporting this work through grants under project IDs a06 and a10. We thank Skander Moalla, Anja Surina, and Lars Quaedvlieg for helpful discussion on SFT tasks. We extend our appreciation to Karin Getaz for administrative support. 
\end{ack}
\bibliographystyle{unsrtnat}
\bibliography{ref}
%%%%%%%%%%%%%%%%%%%%%%%%%%%%%%%%%%%%%%%%%%%%%%%%%%%%%%%%%%%%

% \caption{}
% \label{listing}
% \end{lstlisting}
% \begin{listing}[!t]{algo/gen.py}
% \caption{}
% \label{listing}
% \end{listing}
%%%%%%%%%%%%%%%%%%%%%%%%%%%%%%%%%%%%%%%%%%%%%%%%%%%%%%%%%%%%
\newpage
\section*{NeurIPS Paper Checklist}

\begin{enumerate}

\item {\bf Claims}
    \item[] Question: Do the main claims made in the abstract and introduction accurately reflect the paper's contributions and scope?
    \item[] Answer: \answerYes{} % Replace by \answerYes{}, \answerNo{}, or \answerNA{}.
    \item[] Justification: We ensure that the abstract and introduction clearly summarize the paper's core contributions: the \rat architecture, its intra- and inter-chunk mechanisms, the efficiency, and empirical evaluations on both pretraining and downstream tasks. All major claims are supported by experiments and analyses throughout the paper.

\item {\bf Limitations}
    \item[] Question: Does the paper discuss the limitations of the work performed by the authors?
    \item[] Answer: \answerYes{} % Replace by \answerYes{}, \answerNo{}, or \answerNA{}.
    \item[] Justification: We discuss limitations in \autoref{sec:conclusion}

\item {\bf Theory assumptions and proofs}
    \item[] Question: For each theoretical result, does the paper provide the full set of assumptions and a complete (and correct) proof?
    \item[] Answer: \answerNA{} % Replace by \answerYes{}, \answerNo{}, or \answerNA{}.
    \item[] Justification: The paper does not include theoretical results.

    \item {\bf Experimental result reproducibility}
    \item[] Question: Does the paper fully disclose all the information needed to reproduce the main experimental results of the paper to the extent that it affects the main claims and/or conclusions of the paper (regardless of whether the code and data are provided or not)?
    \item[] Answer: \answerYes{} % Replace by \answerYes{}, \answerNo{}, or \answerNA{}.
    \item[] Justification: We describe our experimental setup and implementation details in \autoref{sec:exp} and \autoref{appendix:sec:impl}.

\item {\bf Open access to data and code}
    \item[] Question: Does the paper provide open access to the data and code, with sufficient instructions to faithfully reproduce the main experimental results, as described in supplemental material?
    \item[] Answer: \answerYes{} % Replace by \answerYes{}, \answerNo{}, or \answerNA{}.
    \item[] Justification: We open-source our code with link put in Appendix.
    The dataset we use is publicly available.

\item {\bf Experimental setting/details}
    \item[] Question: Does the paper specify all the training and test details (e.g., data splits, hyperparameters, how they were chosen, type of optimizer, etc.) necessary to understand the results?
    \item[] Answer: \answerYes{} % Replace by \answerYes{}, \answerNo{}, or \answerNA{}.
    \item[] Justification: We describe our experimental setup in \autoref{sec:exp} and give details in \autoref{appendix:sec:impl}.

\item {\bf Experiment statistical significance}
    \item[] Question: Does the paper report error bars suitably and correctly defined or other appropriate information about the statistical significance of the experiments?
    \item[] Answer: \answerNo{} % Replace by \answerYes{}, \answerNo{}, or \answerNA{}.
    \item[] Justification: We do not report error bars or statistical significance due to computational cost and limited variance observed across repeated runs in preliminary testing. We follow the common practice in similar-scale LLM experiments.

\item {\bf Experiments compute resources}
    \item[] Question: For each experiment, does the paper provide sufficient information on the computer resources (type of compute workers, memory, time of execution) needed to reproduce the experiments?
    \item[] Answer: \answerYes{} % Replace by \answerYes{}, \answerNo{}, or \answerNA{}.
    \item[] Justification: Computing resources are discussed in \autoref{sec:exp} and \autoref{appendix:sec:impl}.
    
\item {\bf Code of ethics}
    \item[] Question: Does the research conducted in the paper conform, in every respect, with the NeurIPS Code of Ethics \url{https://neurips.cc/public/EthicsGuidelines}?
    \item[] Answer: \answerYes{} % Replace by \answerYes{}, \answerNo{}, or \answerNA{}.
    \item[] Justification: The research conducted in the paper conforms with the NeurIPS Code of Ethics.

\item {\bf Broader impacts}
    \item[] Question: Does the paper discuss both potential positive societal impacts and negative societal impacts of the work performed?
    \item[] Answer: \answerYes{} % Replace by \answerYes{}, \answerNo{}, or \answerNA{}.
    \item[] Justification: Societal impacts are discussed in \autoref{appendix:sec:impact}.
    
\item {\bf Safeguards}
    \item[] Question: Does the paper describe safeguards that have been put in place for responsible release of data or models that have a high risk for misuse (e.g., pretrained language models, image generators, or scraped datasets)?
    \item[] Answer: \answerNA{} % Replace by \answerYes{}, \answerNo{}, or \answerNA{}.
    \item[] Justification: We do not include safeguards in the current version, as model and data release is not the focus of this work, and the model is only 1B scale. Should we release models in the future, we will follow community standards to mitigate misuse risk.

\item {\bf Licenses for existing assets}
    \item[] Question: Are the creators or original owners of assets (e.g., code, data, models), used in the paper, properly credited and are the license and terms of use explicitly mentioned and properly respected?
    \item[] Answer: \answerYes{} % Replace by \answerYes{}, \answerNo{}, or \answerNA{}.
    \item[] Justification: All the assets used have been cited.

\item {\bf New assets}
    \item[] Question: Are new assets introduced in the paper well documented and is the documentation provided alongside the assets?
    \item[] Answer: \answerYes{} % Replace by \answerYes{}, \answerNo{}, or \answerNA{}.
    \item[] Justification: We provide documented code.

\item {\bf Crowdsourcing and research with human subjects}
    \item[] Question: For crowdsourcing experiments and research with human subjects, does the paper include the full text of instructions given to participants and screenshots, if applicable, as well as details about compensation (if any)? 
    \item[] Answer: \answerNA{} % Replace by \answerYes{}, \answerNo{}, or \answerNA{}.
    \item[] Justification: The paper does not involve crowdsourcing nor research with human subjects.

\item {\bf Institutional review board (IRB) approvals or equivalent for research with human subjects}
    \item[] Question: Does the paper describe potential risks incurred by study participants, whether such risks were disclosed to the subjects, and whether Institutional Review Board (IRB) approvals (or an equivalent approval/review based on the requirements of your country or institution) were obtained?
    \item[] Answer: \answerNA{} % Replace by \answerYes{}, \answerNo{}, or \answerNA{}.
    \item[] Justification: The paper does not involve crowdsourcing nor research with human subjects.

\item {\bf Declaration of LLM usage}
    \item[] Question: Does the paper describe the usage of LLMs if it is an important, original, or non-standard component of the core methods in this research? Note that if the LLM is used only for writing, editing, or formatting purposes and does not impact the core methodology, scientific rigorousness, or originality of the research, declaration is not required.
    %this research? 
    \item[] Answer: \answerNA{} % Replace by \answerYes{}, \answerNo{}, or \answerNA{}.
    \item[] Justification: LLMs were not used in any component of the core research methodology. Any usage was limited to writing assistance and did not affect the technical contributions of the work.

\end{enumerate}

\clearpage
\definecolor{codegreen}{rgb}{0,0.6,0}
\definecolor{codegray}{rgb}{0.5,0.5,0.5}
\definecolor{codepurple}{rgb}{0.58,0,0.82}
\definecolor{backcolour}{rgb}{0.95,0.95,0.92}

%Code listing style named "mystyle"
\lstdefinestyle{mystyle}{
  backgroundcolor=\color{backcolour}, 
  commentstyle=\color{codegreen}\itshape,
  keywordstyle=\color{magenta},
  numberstyle=\tiny\color{codegray},
  stringstyle=\color{codepurple},
  basicstyle=\ttfamily\footnotesize,
  breakatwhitespace=false,         
  breaklines=true,                 
  captionpos=b,                    
  keepspaces=true,                 
  numbers=left,                    
  numbersep=5pt,                  
  showspaces=false,                
  showstringspaces=false,
  showtabs=false,                  
  tabsize=2
}
\lstset{style=mystyle}
\appendix
\clearpage
\section{Implementation details}
\label{appendix:sec:impl}
\subsection{Algorithm}
\label{appendix:subsec:algo}
We provide the pseudocode for the training and prefilling modes of \rat in \autoref{algo:train_or_prefill}, and the pseudocode for the generation mode in \autoref{algo:gen}. 
% The complete anonymous implementation is put in \url{https://github.com/rat-neurips25/RAT.git}.

\begin{lstlisting}[language=Python, caption={Pseudo code for the training or prefilling modes of \rat. We use Pytorch's \textit{flex attention} and \textit{associative scan} for implementation.}, label={algo:train_or_prefill}]
def merge_last_token_stable(inter_out, intra_out, inter_lse, intra_lse):
    # Merge previous chunk's output and current chunk's output safely
    # by adjusting the base of the exponential in softmax
    max_lse = maximum(inter_lse, intra_lse)
    inter_lse_exp = exp(inter_lse - max_lse)
    intra_lse_exp = exp(intra_lse - max_lse)
    denom = intra_lse_exp + inter_lse_exp
    intra_adjust = (intra_lse_exp / denom).unsqueeze(-1)
    inter_adjust = (inter_lse_exp / denom).unsqueeze(-1)
    return inter_out * inter_adjust + intra_out * intra_adjust


def train_or_prefill(z, g, q, k, v, num_chunk, chunk_size, num_head, softmax_scale):
    """
    Args:
        z: (B, T, D), output gate after sigmoid function
        g: (B, T, D), forget gate after sigmoid function
        v: (B, T, D), value vectors in attention
        q, k: (B, T, D), shared query and key vectors in attention
    Returns:
        out: (B, T, D)
    """
    bs, seq_len, d_model = shape(z)
    k, v = [rearrange(m, "b (c l) d -> b c l d", c=num_chunk) for m in (k, v)]

    # Intra-chunk RNN: apply associative scan along the "l" dimension
    intra_v = ascan(g, (1.0 - g) * v)
    intra_k = ascan(g, (1.0 - g) * k)

    # Rearrange for multi-head attention
    intra_v = rearrange(intra_v, "b c l (a p) -> b a (c l) p", a=num_head)
    intra_k = rearrange(intra_k, "b c l (a p) -> b a (c l) p", a=num_head)
    q = rearrange(q, "b t (a p) -> b a t p", a=num_head)

    # Apply inter-chunk RoPE
    q, intra_k = apply_rope(q, intra_k)

    # Extract chunk's end representation for inter-chunk attention
    chunk_intra_k = intra_k[..., chunk_size - 1::chunk_size, :]
    chunk_intra_v = intra_v[..., chunk_size - 1::chunk_size, :]

    # Inter-chunk attention using flex_attention
    # with chunk-level causal mask and returning log-sum-exp scores
    block_mask = create_block_mask(
        lambda b, h, q_idx, kv_idx: q_idx // chunk_size > kv_idx,
        1, 1, q.shape[2], num_chunk
    )
    inter_out, inter_lse = flex_attention(
        q, chunk_intra_k, chunk_intra_v,
        scale=softmax_scale,
        block_mask=block_mask,
        return_lse=True
    )

    # Compute logits with the current chunk's k, required separately due to causal masking
    intra_lse = einsum("batp, batp -> bat", q, intra_k) * softmax_scale

    # Merge outputs from previous chunks and current chunk (due to causal masking)
    out = merge_last_token_stable(inter_out, intra_v, inter_lse, intra_lse)
    out = rearrange(out, "b a t p -> b t (a p)", a=num_head)
    return z * out
\end{lstlisting}

% \begin{listing}[!ht]
% \caption{Pseudo code for the training or prefilling modes of \rat. We use Pytorch's \textit{flex attention} and \textit{associative scan} for implementation.}
% \inputminted[fontsize=\footnotesize]{python}{algo/train_or_prefill.py}
% \label{algo:train_or_prefill}
% \end{listing}

\begin{lstlisting}[language=Python, caption={Pseudo code for the generation mode of \rat. We simply use the \textit{flash attention} (Pytorch's one). Note that KVCache of \rat is reduced from \(T\) to \(C\) compared to the attention module.}, label={algo:gen}]
def gen(z, g, q, k, v, chunk_start, num_head, cache):
    """
    Args:
        z: (B, 1, D), output gate after sigmoid
        g: (B, 1, D), forget gate after sigmoid
        v: (B, 1, D), value vector
        q, k: (B, 1, D), shared query and key vectors across heads
        chunk_start: int, index of current chunk
        cache: tuple of four tensors
            lastkcache: (B, 1, D), last token k in current chunk (for RNN-style update)
            lastvcache: (B, 1, D), last token v in current chunk (for RNN-style update)
            kcache: (B, A, C, P) or (B, num_head, num_chunk, head_dim): cached k at each chunk's end
            vcache: (B, A, C, P) or (B, num_head, num_chunk, head_dim): cached v at each chunk's end
        # Note: KVCache is stored per chunk, not per token.
    Returns:
        out: (B, 1, D)
    """
    lastkcache, lastvcache, kcache, vcache = cache

    # Intra-chunk RNN: Recurrent update within the current chunk
    intra_k = g * lastkcache + (1.0 - g) * k
    intra_v = g * lastvcache + (1.0 - g) * v
    lastkcache.copy_(intra_k)
    lastvcache.copy_(intra_v)

    # Rearrange input for attention
    intra_k = rearrange(intra_k, "b t (a p) -> b a t p", a=num_head)
    intra_v = rearrange(intra_v, "b t (a p) -> b a t p", a=num_head)
    q = rearrange(q, "b t (a p) -> b a t p", a=num_head)

    # Apply RoPE
    q, intra_k = apply_rope(q, intra_k)

    # Update the current chunk's kvcache
    kcache[:, :, chunk_start:chunk_start + 1] = intra_k
    vcache[:, :, chunk_start:chunk_start + 1] = intra_v

    # Inter-chunk attention: standard attention over cached chunk representations
    attn_out = scaled_dot_product_attention(
        q, kcache[:, :, :chunk_start + 1], vcache[:, :, :chunk_start + 1], is_causal=False
    )
    out = rearrange(out, "b a t p -> b t (a p)", a=num_head)
    return z * out
\end{lstlisting}

% \begin{listing}[!ht]
% \caption{Pseudo code for the generation mode of \rat. We simply use the \textit{flash attention} (Pytorch's one). Note that KVCache of \rat is reduced from \(T\) to \(C\) compared to the attention module.}
% \inputminted[fontsize=\footnotesize]{python}{algo/gen.py}
% \label{algo:gen}
% \end{listing}

\subsection{Experiments}
\label{appendix:subsec:impl}
\paragraph{Dataset} 
In our preliminary studies, we use the PG19 dataset~\citep{pg19}, a long-form English book corpus with inherently long contexts. For the 1.3B model experiments, we adopt the FineWeb-Edu dataset~\citep{fineweb_edu}, using its 100B-token randomly sampled version downloaded from the HuggingFace repository. To match the pretraining context length, we concatenate documents using a separator token.  Note that web samples are usually very short, compared to the book dataset. For downstream evaluation, we consider a suite of classical commonsense reasoning benchmarks from the Eleuther AI evaluation harness~\citep{eval-harness}, including PIQA~\citep{piqa}, ARC-C~\citep{arc}, and HellaSwag~\citep{hellaswag}. In the LLaMA2 tokenizer, we observe that the inputs of these tasks typically contain fewer than 300 tokens. For the LongBench evaluation, in \autoref{tab:longbench} we provide input length for each task, offering a rough indication of task difficulty with respect to input length. For SFT-based tasks, we have elaborated sequence lengths in the main text. As LongBench and SFT tasks often involve very long inputs, we apply truncation in the middle to preserve information at both the beginning and the end.
\begin{table}[htbp]
\centering
\caption{We report the average input length and the 95th percentile input length of each task, measured in tokens using the LLaMA2 tokenizer. NQA (NarrativeQA), MQA (MultiFieldQA-en), HQA (HotpotQA), WQA (2WikiMultihopQA), MSQ (Musique), GR (GovReport), MN (MultiNews), TQA (TriviaQA), and RBP (RepoBench-P). 
}
\resizebox{\textwidth}{!}{%
\begin{tabular}{l ccc ccc ccc cc}
\toprule
\bf {Task}
& \multicolumn{3}{c}{\textbf{Single-Document QA}} 
& \multicolumn{3}{c}{\textbf{Multi-Document QA}} 
& \multicolumn{3}{c}{\textbf{Summarization}} 
& \multicolumn{2}{c}{\textbf{Code Completion}} \\
% & \bf NQA & \bf Qasper & \bf MQA & \bf Avg.
% & \bf HQA & \bf WQA & \bf MSQ & \bf Avg.
% & \bf GR & \bf QMSum  & \bf \bf MN & \bf Avg.
% & \bf TREC & \bf TQA & \bf SAMSum & \bf Avg.
% & \bf LCC & \bf RBP & \bf Avg. \\
% & F1 & F1 & F1 & -
% & F1 & F1 & F1 & -
% & Rouge-L & Rouge-L  & Rouge-L& -
% & Acc & F1 & Rouge-L & -
% & EditSum & EditSum & -\\
\midrule
\bf Name & NQA & Qasper & MQA & HQA & WQA & MSQ & GR & QMSum & MN & LCC & RBP \\
\midrule
\bf Avg. & 36037 & 5780 & 8115 & 15329 & 8483 & 18555 & 12280 & 15980 & 3156 & 4307 &  14818 \\
\bf 95th pctl. &  77966 & 10164 & 14994 & 19755 & 16939 & 20066 & 25721 & 29069 & 7135 & 10401 & 31937 \\
\bottomrule
\label{tab:longbench}
\end{tabular}%
}
\end{table}

\paragraph{200M model}
We start with a 200M-parameter model in our preliminary study, with a model dimension of 1024, 12 transformer layers, and head dimension of 64. The rotary position embedding (RoPE) base is set to 10{,}000. We use the GPT2 tokenizer. Following~\citet{landmarkattention}, we repeat the PG19 training split five times to reach a total of 15B training tokens. The learning rate is scheduled using cosine annealing, starting at \(6.0 \times 10^{-4}\) and decaying to \(1.0 \times 10^{-6}\), with a warm-up ratio of 10\%. We use the AdamW optimizer with a weight decay of 0.1 and \(\beta = (0.9, 0.98)\). Gradient clipping is applied with a threshold of 1.0, and the global batch size is set to 1M tokens. We explore training with three different context lengths: 8K, 16K, and 32K. Training these models on 4 H100 GPUs takes approximately 5 to 14 hours. In particular, the attention model requires up to 14 hours when the sequence length is \(T = 32768\), whereas the \ratl{16} model completes training in about 7 hours under the same setting.

\paragraph{1.3B model}
The 1.3B-parameter model uses a model dimension of 2048, 24 transformer layers, and a head dimension of 128, equipped with RMSNorm~\citep{rmsnorm} and without bias. The RoPE base is also set to 10{,}000. The model parameters are initialized using a Gaussian distribution with a standard deviation of 0.02. We adopt the LLaMA2 tokenizer in the following studies. For pretraining, we use a cosine-annealed learning rate schedule starting at \(8.0 \times 10^{-4}\) and decaying to \(1.0 \times 10^{-6}\), with 5\% warmup. The global batch size is set to 2M tokens, and the context window is set to 4096. Each model is trained on 16 H100 GPUs, requiring approximately 2 to 3 days to complete.

For LongBench evaluation, we follow the default prompts with greedy decoding for all tasks except summarization. For summarization, we apply a repetition penalty of 1.2 to address the common issue in pretrained-only models of generating repetitive outputs under instructional prompts. For SFT tasks, we train the models on the official training splits with an answer-only loss and evaluate them on the corresponding test sets. Although we explored different hyperparameters during the early experimentation, we observed that the relative trends remained largely stable. Thus, we fix the learning rate and batch size to \((1.0 \times 10^{-5}, 128)\) for large datasets, and \((1.0 \times 10^{-5}, 32)\) for the smaller QMSum~\citep{qmsum} task, following common practice for 1B-scale models. The weight decay is set as 0.01, and all other hyperparameters follow the pretraining setup. We sweep over the number of epochs and report the best result for each dataset and architecture. In practice, QA tasks typically converge in 1–2 epochs, while summarization benefits from slightly longer training.
\section{Efficiency}
\label{appendix:sec:efficiency}
\subsection{Latency}
\label{appendix:subsec:latency}
To supplement \autoref{fig:eff} in main text, we put the concrete latency number of a single layer in \autoref{tab:eff_train}, \autoref{tab:eff_prefill}, and \autoref{tab:eff_gen}.

\begin{table}[htbp!]
\centering
\caption{Single temporal-mixing layer (including input and output projections) training time across different sequence lengths. The latency (ms) is tested on 200K tokens.}
\begin{adjustbox}{max width=0.9\textwidth}
\begin{tabular}{lrrrrrrrr}
\toprule
Model & 4096 & 8192 & 16384 & 32768 & 65536 & 131072 & 262144 \\
\midrule
RNN & 77.00 & 78.42 & 77.50 & 83.50 & 115.83 & 130.62 & 195.30 \\
\bf \ratl{128} & 150.19 & 150.76 & 156.65 & 154.52 & 159.94 & 165.07 & 206.46 \\
\bf \ratl{64} & 146.25 & 148.46 & 151.10 & 155.83 & 165.58 & 177.78 & 227.76 \\
\bf \ratl{16} & 150.08 & 153.63 & 158.98 & 173.89 & 202.00 & 249.79 & 378.11 \\
Attention & 122.06 & 176.44 & 277.61 & 474.90 & 877.14 & 1713.82 & 3417.48 \\
\midrule
Attention/\ratl{16} & 0.81\(\times\) & 1.15\(\times\) & 1.75\(\times\) & 2.73\(\times\) & 4.34\(\times\) & 6.86\(\times\) & 9.04\(\times\) \\
\bottomrule
\end{tabular}
\end{adjustbox}
\label{tab:eff_train}
\end{table}

\begin{table}[htbp!]
\centering
\caption{Single temporal-mixing layer (including input and output projections) prefilling time across different sequences lengths. The latency (ms) is tested on 200K tokens.}
\begin{adjustbox}{max width=0.9\textwidth}
\begin{tabular}{lrrrrrrrr}
\toprule
Model & 4096 & 8192 & 16384 & 32768 & 65536 & 131072 & 262144 \\
\midrule
RNN & 24.93 & 24.75 & 25.32 & 27.09 & 30.54 & 39.40 & 56.16 \\
\ratl{128} & 44.74 & 44.78 & 45.09 & 45.50 & 46.67 & 48.62 & 51.75 \\
\ratl{64} & 43.46 & 44.27 & 44.27 & 44.83 & 47.08 & 50.83 & 57.38 \\
\ratl{16} & 44.03 & 44.90 & 45.53 & 51.08 & 57.31 & 71.18 & 99.53 \\
Attention & 36.93 & 52.71 & 80.21 & 135.68 & 245.14 & 494.62 & 997.50 \\
\midrule
Attention/\ratl{16} & 0.84\(\times\) & 1.17\(\times\) & 1.76\(\times\) & 2.66\(\times\) & 4.28\(\times\) & 6.95\(\times\) & 10.02\(\times\) \\
\bottomrule
\end{tabular}
\end{adjustbox}
\label{tab:eff_prefill}
\end{table}

\begin{figure}[htbp!]
    \centering
    \includegraphics[width=0.95\textwidth]{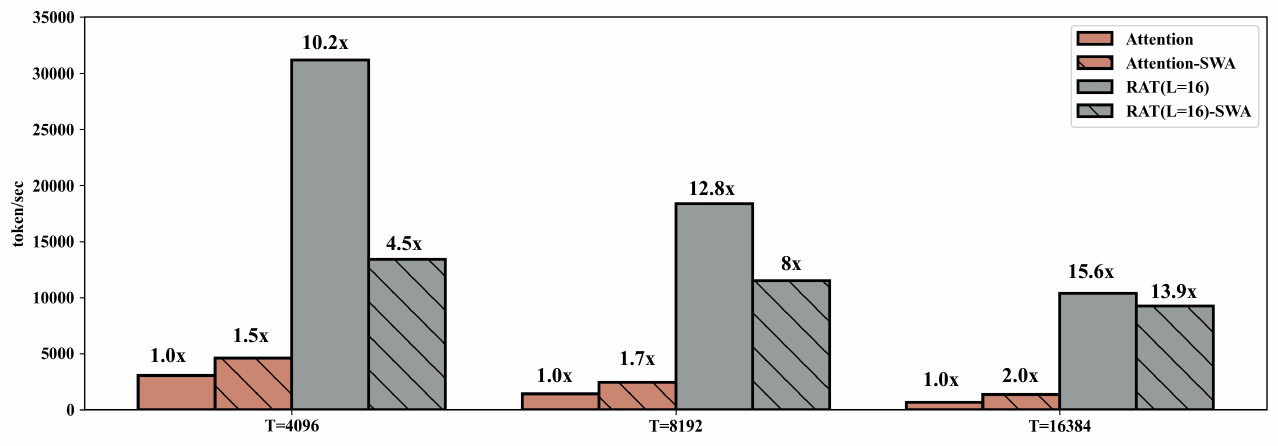}
    \caption{We measure the maximum throughput of the full 1.3B model for generating 1024 tokens under different prefilling lengths. For each total sequence length \(T\), the prefilling length is set to \(T - 1024\). For example, \(T=4096\) corresponds to a prefilling of 3072 tokens, while \(T=8192\) and \(T=16384\) correspond to 7168 and 15360 tokens, respectively. When the sequence length increases, the maximum throughput ratio between \rat and attention rises from \(10.2\times\) to \(15.6\times\), highlighting the strong efficiency advantage of \rat in long-context generation.}
\label{fig:throughput}
\end{figure}

\begin{table}[htbp!]
\centering
\caption{Single temporal-mixing layer (including input and output projections) generation time at the specified position. The latency (ms) is tested on generating batches of tokens with \(B=64\), \(B=512\), and \(B=1024\).}
\begin{adjustbox}{max width=0.9\textwidth}
\begin{tabular}{lrrrrrrrr}
\toprule
Model & 4096 & 8192 & 16384 & 32768 & 65536 & 131072 & 262144 \\
\midrule
\(B=64\) \\
\midrule
\hspace{0.5em}RNN & 0.36 & 0.33 & 0.34 & 0.33 & 0.33 & 0.33 & 0.33 \\
\hspace{0.5em}\ratl{128} & 0.62 & 0.60 & 0.61 & 0.63 & 0.66 & 0.66 & 0.68 \\
\hspace{0.5em}\ratl{64} & 0.67 & 0.64 & 0.66 & 0.70 & 0.73 & 0.74 & 1.03\\
\hspace{0.5em}\ratl{16} & 0.62 & 0.65 & 0.64 & 0.68 & 1.04 & 1.82 & 3.42 \\
\hspace{0.5em}Attention & 1.15 & 1.73 & 3.33 & 6.66 & 12.94 & 26.8 & OOM \\
\midrule
\(B=512\) \\
\midrule
\hspace{0.5em}RNN & 0.35 & 0.34 & 0.33 & 0.33 & 0.34 & 0.34 & 0.33 \\
\hspace{0.5em}\ratl{128} & 0.63 & 0.64 & 0.66 & 0.75 & 1.17 & 2.44 & 3.46 \\
\hspace{0.5em}\ratl{64} & 0.70 & 0.68 & 0.74 & 1.16 & 2.41 & 3.44 & 6.69 \\
\hspace{0.5em}\ratl{16} & 0.75 & 1.22 & 2.45 & 3.48 & 6.73 & 13.09 & 25.76 \\
\hspace{0.5em}Attention & 6.56 & 12.89 & 25.84 & OOM & OOM & OOM & OOM \\
\midrule
\(B=1024\) \\
\midrule
\hspace{0.5em}RNN & 0.36 & 0.36 & 0.35 & 0.35 & 0.35 & 0.36 & 0.35 \\
\hspace{0.5em}\ratl{128} & 0.73 & 0.75 & 0.95 & 1.35 & 2.54 & 4.64 & 6.98 \\
\hspace{0.5em}\ratl{64} & 0.74 & 0.94 & 1.35 & 2.48 & 4.67 & 7.02 & 13.45\\
\hspace{0.5em}\ratl{16} & 1.38 & 2.56 & 4.63 & 7.04 & 13.51 & 26.43 & OOM \\
\hspace{0.5em}Attention & 13.04 & 25.70 & OOM & OOM & OOM & OOM & OOM \\
\bottomrule
\end{tabular}
\end{adjustbox}
\label{tab:eff_gen}
\end{table}

\subsection{Maximum throughput}
\label{appendix:subsec:throughput}
As \rat also reduces cache memory usage, we report the maximum throughput of the full 1.3B model across different sequence lengths in \autoref{fig:throughput}. For example, \ratl{16} achieves an approximately \(10.2\times\) higher throughput than the baseline attention model at \(T = 4096\), and improves further to \(15.6\times\) at \(T = 16384\). Similarly, \ratlswa{16} reaches \(4.5\times\), \(8.0\times\), and \(13.9\times\) higher throughput over attention at \(T = 4096\), \(8192\), and \(16384\), respectively.

\section{Accuracy}
\label{appendix:sec:accuracy}

\subsection{Supplementary ablation study}
All experiments here are conducted on the PG19 Book dataset with 200M-parameter models, and FLOPs are controlled for comparison.
\paragraph{Reversed design of \rat}
\label{appendix:subsec:ablation}
Instead of employing recurrence within each chunk, we also explored a reversed design that applies attention inside the chunk, followed by recurrence to capture long-range dependencies. For this reversed design, standard RoPE is sufficient. Regarding the parameter allocation, we found it crucial to retain the key vector in the attention module as a full tensor, as sharing it together with the query vector leads to a collapse into single-head attention. Thus, the final reversed design shares the query vector across attention heads, employs low-rank matrices for the projection used for output gate, and preserves full projections for both the forget gate and the key vector. 

Results of the reversed design are put in \autoref{tab:reversed_design}. It can be observed that, under identical FLOP constraints, this design significantly underperforms \rat. We think this is because short-context dependencies are easier to capture and do not suffer from memory degradation, so using lightweight recurrence locally is more efficient. Long‑context dependencies are harder and may require direct retrieval of distant information, where attention is more suitable. For example, to match the FLOPs budget of \(\mathcal{O}(T/64)\), the reversed design requires chunk sizes of 128, 256, and 512 for sequence lengths of 8192, 16384, and 32768, respectively. With this small local window size, the long-range dependencies are attributed to the recurrence, thus leading to high perplexity. 
Only when we reduce the number of chunks to 16 does the perplexity drop below 14.00. Therefore, while the reversed design also allows for interpolation between attention and RNN (reducing to attention when \(L = T\), and to RNN when \(L = 1\)), we choose to focus on \rat due to its more efficient utilization of FLOPs.

\paragraph{Comparison to FLASH~\citep{gau}}
As pointed out in \autoref{sec:related_work}, \citet{gau} also uses the concept of chunking; however, we adopt a fundamentally different framework, both in the components involved and in how the two outputs are combined. Specifically, \citet{gau} simply adds the outputs of inter-chunk and intra-chunk computations, whereas \rat organizes them hierarchically, applying recurrence over the key and value before inter-chunk attention. We argue that these differences lead to the weaker performance of \citet{gau}, as shown in \autoref{tab:flash}. First, as discussed above, \rat utilizes FLOPs more effectively than methods based on local attention, since its long-range dependencies must rely on memory-degrading components such as recurrence and linear attention. Second, we find that directly adding the outputs of softmax-based attention and linear attention introduces differences in output scale and potential representational conflicts, which may lower down the performance.

\paragraph{Comparison to dilated attention}
We discuss the difference between \rat and the design that interleaves recurrence and dilated attention~\citep{longformer} with the sliding pattern. First, from an efficiency perspective, RAT has advantages over dilated attention in both training and inference. During training, dilated attention reduces parallelism by the dilation rate, since tokens within the same dilation span attend to different KV vectors. In contrast, RAT allows tokens to share the same chunk-level representation, except within their own chunk. During inference, the sliding dilation pattern causes a memory issue, since the KV cache of all tokens must be stored, whereas RAT reduces the cache size proportionally to the chunk size. Second, in terms of accuracy, to match the FLOPs, we set the dilation rate to 64 for RNN-Dilated attention, since dilated attention occurs in only half of the layers. It can be seen from \autoref{tab:dilated_attention} that dilated attention performs poorly. We observed significantly slower convergence at the beginning of training, likely due to its lack of global perception.

\begin{minipage}[t]{1.0\textwidth}
\begin{minipage}[t]{0.5\textwidth}
\centering
\captionof{table}{Perplexity results of the reversed design, where attention is applied within chunks and RNN is used across chunks. Experiments are conducted on the 200M model trained on the PG19 book dataset. Under the same FLOPs, it can be observed that \rat significantly outperforms its reversed counterpart.}
\label{tab:reversed_design}
\begin{adjustbox}{max width=\textwidth}
\begin{tabular}{ll cccc}
\toprule
\bf Method & \bf FLOPs & \(\mathbf{T=8192}\) & \(\mathbf{T=16384}\) & \(\mathbf{T=32768}\)\\
\midrule
\ratl{128}  & \(\mathcal{O}(T/128)\) & 13.57 & 13.44 & 13.66\\
\ratl{64} & \(\mathcal{O}(T/64)\) & 13.42 & 13.34 & 13.39 \\
\midrule
\texttt{RAT-Reversed(C=128)} & \(\mathcal{O}(T/128)\) & 14.53 & 14.35 & 14.37 \\
\texttt{RAT-Reversed(C=64)} & \(\mathcal{O}(T/64)\) & 14.21 & 14.06 & 14.02 \\
\texttt{RAT-Reversed(C=16)} & \(\mathcal{O}(T/16)\) & 13.69 & 13.48 & 13.50 \\
\midrule
Attention & \(\mathcal{O}(T)\) & 13.26 & 13.10 & 13.19 \\
\bottomrule
\end{tabular}
\end{adjustbox}
\end{minipage}
\hfill
\begin{minipage}[t]{0.48\textwidth}
\centering
\captionof{table}{Pretraining and downstream evaluation results of 1.3B models using either RoPE or NoPE positional encodings. RoPE is used in the main text, while NoPE is trained for investigating length extrapolation. Notably, NoPE achieves reasonable performance even at the 1B model scale with only 100B training tokens.}
\label{tab:nope}
\begin{adjustbox}{max width=\textwidth}
\begin{tabular}{ll cccc}
\toprule
\bf Method & \bf PE & \bf Pretrain & \bf HellaSwag & \bf LAMBADA & \bf PIQA\\
& & PPL & acc\_norm & acc & acc\_norm \\
\midrule
Attention-SWA & RoPE & 7.61 & 57.1 & 48.0 & 72.7\\
\ratlswa{16} & RoPE & 7.57 & 57.9 & 49.3 & 73.9\\
\midrule
Attention-SWA & NoPE & 7.69 & 56.4 & 47.3 & 72.7\\
\ratlswa{16} & NoPE &  7.63 & 56.8 & 47.8 & 73.7\\
\bottomrule
\end{tabular}
\end{adjustbox}
\end{minipage}
\end{minipage}

\begin{minipage}[t]{1.0\textwidth}
\begin{minipage}[t]{0.67\textwidth}
\centering
\captionof{table}{Performance of different chunk-based designs with \(T=16384\) and FLOPs \(\mathcal{O}(T/64)\).}
\label{tab:flash}
\begin{adjustbox}{max width=\textwidth}
\begin{tabular}{lllll}
\toprule
\bf Method & \bf Intra-chunk & \bf Inter-chunk & \bf Organization & \bf PPL\\
\midrule
\ratl{64} & Recurrence	& Attention	& Hierarchically & \bf 13.34\\
\texttt{RAT-Reversed(C=64)} & Attention	& Recurrence	& Hierarchically 
& 14.06 \\
FLASH (C=128) & Attention	& Linear attention	& Add together & 14.4 \\
\bottomrule
\end{tabular}
\end{adjustbox}
\end{minipage}
\hfill
\begin{minipage}[t]{0.31\textwidth}
\centering
\captionof{table}{Performance compared to dilated attention with sliding patterns under the same FLOPs.}
\label{tab:dilated_attention}
\begin{adjustbox}{max width=\textwidth}
\begin{tabular}{ll}
\toprule
\bf Method & PPL\\
\midrule
RNN-Dilated attention & 15.22 \\
\ratl{128} & 13.44 \\
\bottomrule
\end{tabular}
\end{adjustbox}
\end{minipage}
\end{minipage}

\subsection{Length generalization}
\label{appendix:subsec:generalization}
\begin{figure}[htbp]
    \centering
    \includegraphics[width=0.95\textwidth]{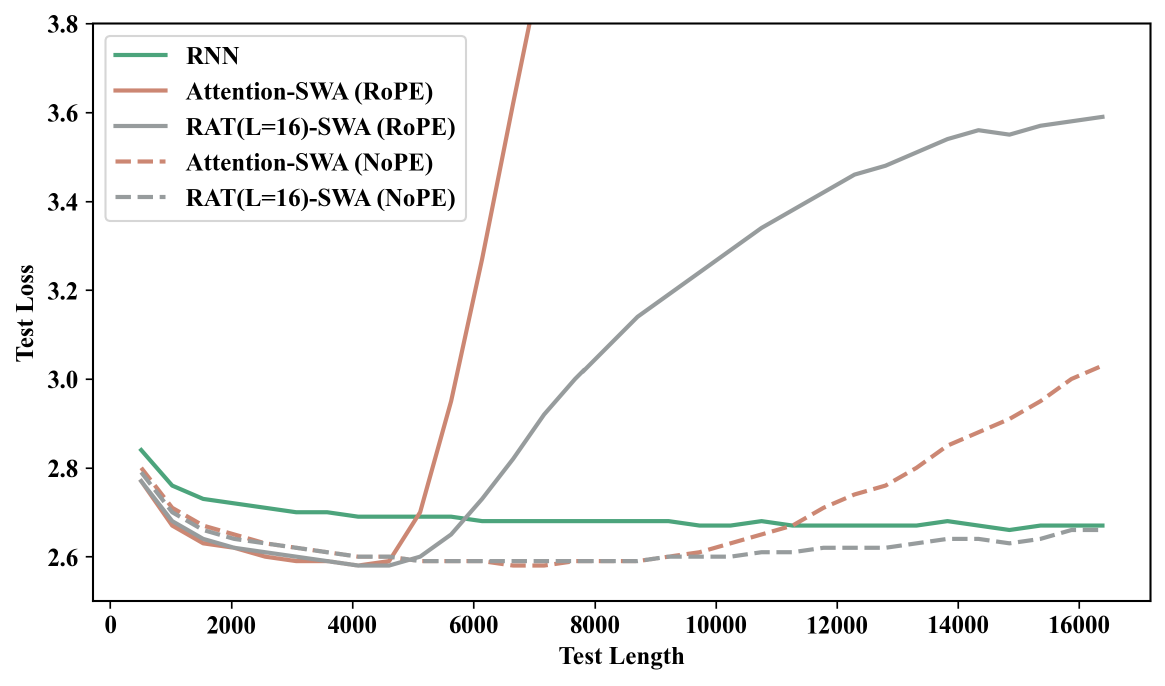}
    \caption{Evaluation at different test lengths for pretrained models trained with a 4K context window. \ratlswa{16} with NoPE achieves the best overall performance, exhibiting strong generalization up to \(T = 16384\) while maintaining low loss within the training context.}
\label{fig:generalization}
\end{figure}
Because of the use of softmax-based attention and RoPE at the inter-chunk level, it is reasonable to expect that \rat may also face challenges in length generalization, where the model is trained on short sequences but evaluated on much longer ones. To study this, we consider SWA variants using either RoPE or NoPE for attention and \rat, following recent practices that interleave attention with local attention modules and apply positional encodings only in the local attention. This design has been adopted in practice and has shown promising results~\citep{cohere_llm}.

As shown in \autoref{fig:generalization}, the RNN performs very steadily as the test sequence length increases. However, within the training context, it has the highest loss among all models. With RoPE, both Attention-SWA and \ratlswa{16} experience a sharp increase in loss when the sequence length goes beyond 6000. The increase is more severe in the attention model than in the \rat variant. This is likely because \rat reduces the effective attention span by attending only to inter-chunk positions, which may offer better robustness to extrapolation.

We also pretrain NoPE versions, where no positional encodings are used in attention or \rat layers. Interestingly, these NoPE models show reasonable performance, as reported in \autoref{tab:nope}. While their pretraining perplexities are higher than those of the RoPE variants, the gap remains small, especially considering the 1.3B model size and 100B-token training budget. To further verify this, we evaluate the models on several commonsense reasoning tasks and obtain promising results. We expect that NoPE will perform even better with larger models and longer training. In terms of length extrapolation, using NoPE significantly improves the robustness for both Attention-SWA and \ratlswa{16}. Among them, \ratlswa{16} with NoPE shows the most stable performance, maintaining low loss even at a sequence length of 16,384.

In conclusion, we find that \rat can outperform the attention module in length generalization, both with RoPE and with NoPE. Still, the problem is not fully solved, as none of the models reach the stability level of the simple RNN. While many techniques have been proposed to improve length extrapolation in attention models, they should also be considered for \rat, as \rat essentially redirects attention from every position to inter-chunk locations. For example, RoPE extrapolation in the attention module can be improved using methods from \citet{yarn,rope_ntk}, and NoPE extrapolation has been studied in \citet{nope_head_scale}, which points out that NoPE still has a context length limit, though it performs better than RoPE. Their work links the failure to shifts in attention distributions and proposes tuning the temperature of attention heads to improve extrapolation. We believe such methods could also be applied to \rat, and leave further exploration of these techniques to future work.

\subsection{Needle-in-Haystack (retrieval ability)}
\label{appendix:subsec:evaluation}
To evaluate retrieval capabilities---an area where attention-based architectures are known to excel, and where RNNs typically fall short---we conduct experiments on the Needle-in-Haystack tasks. In these tasks, a "needle" (e.g., a magic number or UUID) is embedded within irrelevant passages, noisy text, or mixed with multiple key-value pairs. Each key-value pair is presented as a short identifier (the key) followed by its associated value, and the model is prompted to retrieve the correct value given a specific key.

We adopt the RULER benchmark~\citep{ruler} and evaluate a range of Needle-in-Haystack task configurations. Specifically, task \texttt{single\_1} involves retrieving a specific number from a background filled with repeated noise. In \texttt{single\_2}, the background is replaced with natural stories. \texttt{single\_3} increases the difficulty by requiring retrieval of a long and complex UUID. The \texttt{multikey} tasks are more challenging than the single-key ones, as they introduce multiple such key-value pairs into the context. In particular, \texttt{multikey\_2} and \texttt{multikey\_3} consist almost entirely of densely packed key-value pairs, among which only a single key is queried at the end. This makes the task especially challenging, as the model must retrieve the correct item from a highly cluttered input full of distractors. And \texttt{multikey\_3} further incorporates the complex UUID format. In the \texttt{multiquery} and \texttt{multivalue} settings, the model is required to resolve multiple retrieval targets, either by answering several distinct queries or by retrieving all values associated with a single key. For detailed definitions of each task, we refer the reader to the benchmark documentation.

During evaluation, we observed that models may fail to interpret certain prompts correctly. For instance, prompts like \emph{"A special magic number is hidden within the following text. Make sure to memorize it. I will quiz you about the number afterwards."} can cause failures even in attention-based models, especially on the harder \texttt{multikey} tasks. To mitigate this, we apply a light, one-round supervised fine-tuning stage before evaluation to adapt the models to the instruction patterns. We generate 1000 synthetic training samples for each of the 8 tasks, resulting in a total of 8000 examples disjoint from the validation sets. The models are trained on this dataset for one round and then are evaluated directly on all 8 tasks. This procedure yields a fairer and more stable comparison. As shown in \autoref{tab:niah}, RNNs perform reasonably well on the simpler tasks (\texttt{single\_1}, \texttt{single\_2}), but their performance drops to near zero on the harder ones. Attention-based models perform consistently well, benefiting from their ability to access all previous tokens directly. \ratl{16} achieves results close to full attention, especially in numeric tasks. However, it struggles on UUID tasks due to their complexity and the use of exact match scoring (\texttt{single\_3}, \texttt{multikey\_3}). Meanwhile, \ratl{64} falls between RNNs and \ratl{16}, as expected given its partial access to long-range context.

These results are consistent with the underlying architecture designs: attention provides full direct access to all past tokens, RNNs compress all past information into a hidden state, while \rat compresses part of the history but also retains direct access at the chunk level. As a result, \rat naturally exhibits retrieval capabilities that lie between those of attention and RNNs.

\section{Broader Impacts}
\label{appendix:sec:impact}
Enhancing the efficiency of Large Language Models (LLMs) can significantly reduce computational resources and energy consumption, benefiting the environment and democratizing access to advanced AI technologies.
However, increased efficiency could also lead to greater dissemination of disinformation and the creation of deepfakes, posing risks to public trust and security and potentially reinforcing existing biases that impact specific groups unfairly.
This research aims to promote the responsible development and deployment of LLMs, maximizing societal benefits while acknowledging potential harms.

\section{License information}

\begin{itemize}
    \item FineWeb-Edu (dataset): Open Data Commons License Attribution family. 

    Link: \url{https://huggingface.co/datasets/HuggingFaceFW/fineweb-edu}
    
    \item LongBench (dataset and code): MIT License. 

    Link: \url{https://github.com/THUDM/LongBench}
    
    \item NarrativeQA (dataset): Apache License 2.0.

    Link: \url{https://github.com/deepmind/narrativeqa}
    
    \item QMSum (dataset): MIT License.
    
    Link: \url{https://github.com/Yale-LILY/QMSum}
    
    \item WikiSum (dataset): Custom license (unspecified). 
    
    Link: \url{https://huggingface.co/datasets/d0rj/wikisum}
    
    \item lm-evaluation-harness (code): MIT License.
    
    Link: \url{https://github.com/EleutherAI/lm-evaluation-harness}
    
    \item RULER benchmark (code): Apache 2.0 License. 

    Link: \url{https://github.com/NVIDIA/RULER}
\end{itemize}

\end{document}